\documentclass[sigconf]{acmart}

\usepackage{graphicx}
\usepackage{balance}  

\usepackage{caption}
\usepackage{subcaption}
\usepackage{soul}
\usepackage[linesnumbered,ruled]{algorithm2e}
\usepackage{algorithmic}
\usepackage{ifthen}
\usepackage{amsfonts}

\usepackage{booktabs} 
\usepackage{enumitem}
\usepackage{comment}

\usepackage{hyperref}

\usepackage{color, colortbl}
\definecolor{Gray}{gray}{0.9}

\usepackage{multirow}

\theoremstyle{definition}
\newtheorem{definition}{Definition}[section]

\usepackage{relsize}

\usepackage{slashbox}

\usepackage{setspace}
\usepackage{microtype}

\usepackage{xcolor,colortbl}

\definecolor{Gray}{gray}{0.70}
\definecolor{Gray2}{gray}{0.90}
\definecolor{LightCyan}{rgb}{0.88,1,1}

\newcolumntype{b}{>{\columncolor{Gray}}c}
\newcolumntype{a}{>{\columncolor{Gray2}}c}
\newcolumntype{d}{>{\columncolor{LightCyan}}c}

\setcopyright{acmcopyright}

\begin{document}
\setlength{\pdfpagewidth}{8.5in}
\setlength{\pdfpageheight}{11in}

\title[S. Moosavi et al.]{Accident Risk Prediction based on Heterogeneous Sparse Data: New Dataset and Insights}

\author[ ]{Sobhan Moosavi}
\affiliation{%
  \institution{The Ohio State University}
  \streetaddress{2015 Neil Ave}
  \city{Columbus}
  \state{Ohio}
  \postcode{43210}
}
\email{moosavi.3@osu.edu}

\author[ ]{Mohammad Hossein Samavatian}
\affiliation{%
  \institution{The Ohio State University}
  \streetaddress{2015 Neil Ave}
  \city{Columbus}
  \state{Ohio}
  \postcode{43210}
}
\email{samavatian.1@osu.edu}

\author[ ]{Srinivasan Parthasarathy}
\affiliation{%
  \institution{The Ohio State University}
  \streetaddress{2015 Neil Ave}
  \city{Columbus}
  \state{Ohio}
  \postcode{43210}
}
\email{srini@cse.ohio-state.edu}

\author[ ]{Radu Teodorescu}
\affiliation{%
  \institution{The Ohio State University}
  \streetaddress{2015 Neil Ave}
  \city{Columbus}
  \state{Ohio}
  \postcode{43210}
}
\email{teodores@cse.ohio-state.edu}

\author[ ]{Rajiv Ramnath}
\affiliation{%
  \institution{The Ohio State University}
  \streetaddress{2015 Neil Ave}
  \city{Columbus}
  \state{Ohio}
  \postcode{43210}
}
\email{ramnath@cse.ohio-state.edu}

\begin{abstract}
Reducing traffic accidents is an important public safety challenge, therefore, accident analysis and prediction has been a topic of much research over the past few decades. Using small-scale datasets with limited coverage, being dependent on extensive set of data, and being not applicable for real-time purposes are the important shortcomings of the existing studies. To address these challenges, we propose a new solution for real-time traffic accident prediction using easy-to-obtain, but sparse data. Our solution relies on a deep-neural-network model (which we have named {\it DAP}, for Deep Accident Prediction); which utilizes a variety of data attributes such as {\it traffic events}, {\it weather data}, {\it points-of-interest}, and {\it time}. DAP incorporates multiple components including a recurrent (for time-sensitive data), a fully connected (for time-insensitive data), and a trainable embedding component (to capture spatial heterogeneity). To fill the data gap, we have - through a comprehensive process of data collection, integration, and augmentation - created a large-scale publicly available database of accident information named \textit{US-Accidents}. By employing the US-Accidents dataset and through an extensive set of experiments across several large cities, we have evaluated our proposal against several baselines. Our analysis and results show significant improvements to predict rare accident events. Further, we have shown the impact of traffic information, time, and points-of-interest data for real-time accident prediction. 
\end{abstract}



\copyrightyear{2019}
\acmYear{2019}
\acmConference[SIGSPATIAL '19]{27th ACM SIGSPATIAL International Conference on Advances in Geographic Information Systems}{November 5--8, 2019}{Chicago, IL, USA}
\acmBooktitle{27th ACM SIGSPATIAL International Conference on Advances in Geographic Information Systems (SIGSPATIAL '19), November 5--8, 2019, Chicago, IL, USA}
\acmPrice{15.00}
\acmDOI{10.1145/3347146.3359078}
\acmISBN{978-1-4503-6909-1/19/11}

\maketitle

\section{Introduction}
\label{sec:intro}
Reducing traffic accidents is an important public safety challenge around the world. A global status report on traffic safety \cite{world2015global} notes that there were 1.25 million traffic deaths in 2013 alone, with deaths increasing in 68 countries when compared to 2010. Accident prediction is important for optimizing public transportation, enabling safer routes, and cost-effectively improving the transportation infrastructure, all in order to make the roads safer. Given its significance, accident analysis and prediction has been a topic of much research in the past few decades. Analyzing the impact of environmental stimuli (e.g., road-network properties, weather, and traffic) on traffic accident occurrence patterns \cite{eisenberg2004mixed,jaroszweski2014influence,tamerius2016precipitation}, predicting frequency of accidents within a geographical region \cite{chang2005data,caliendo2007crash,najjar2017combining,ren2018deep,yuan2018hetero}, and predicting risk of accidents \cite{wenqi2017model,yuan2017predicting,lin2015novel,chen2016learning} are the major related research categories. 

Employing small-scaled datasets with limited coverage (e.g. a small number of road-segments, or just one city) \cite{chang2005analysis,chang2005data,caliendo2007crash,lin2015novel,wenqi2017model}; being dependent on a wide range of data attributes which may not be available for all regions (e.g., satellite imagery, traffic volume, and properties of road-network) \cite{yuan2017predicting,najjar2017combining,yuan2018hetero}; being not applicable for real-time applications regarding the modeling constraints and prerequisites (e.g., prediction for longer time intervals such as one day or one week, or requiring extensive set of data) \cite{caliendo2007crash,ren2018deep,yuan2018hetero,najjar2017combining}; and employing over simplified methods for traffic accident prediction \cite{lin2015novel,ihueze2018road,caliendo2007crash} are the main shortcomings of the existing studies. 

To address these challenges and provide a reasonable solution for real-time traffic accident prediction, we propose {\it DAP}, a deep-neural-network-based accident prediction model. DAP uses a variety of data including {\em traffic events} (e.g., congestion, construction, and road hazards), {\em weather} (e.g., temperature, visibility, and wind speed), {\em points-of-interest} (e.g., traffic signal, stop sign, and junction), and {\em time} (e.g., day of week, hour of day, and period of day) to provide real-time prediction for a geographical region of reasonable size (i.e., a square of size $5km\, \times 5km$ on map) and during a fine-grained time period (i.e., a 15 minutes interval). To our knowledge, this is the first research work that has employed traffic events and points-of-interest data for accident prediction. DAP exerts multiple important components to utilize different categories of attributes. To utilize time-sensitive data (e.g., traffic, weather, and time data), DAP employs a recurrent component with Long Short Term Memory (LSTM) cells. To utilize time-insensitive data (e.g., points-of-interest), DAP employs feed-forward neural network layers. Further, to better capture spatial heterogeneity, which has been proven to be effective for accident prediction \cite{yuan2017predicting}, DAP employs trainable latent representation for each geographical region to encode essential spatiotemporal information. 

In order to mitigate the impact of data size on analysis and prediction, we present a new dataset, we name it {\em US-Accidents}, which includes about $2.25$ million traffic accidents took place within the contiguous United States\footnote{The contiguous United States excludes Alaska and Hawaii, and considers District of Columbia (DC) as a separate state.}, between February 2016 and March 2019. US-Accidents offers a wide range of data attributes to describe each accident including {\em location data}, {\em time data}, {\em natural language description of event}, {\em weather data}, {\em period-of-day information}\footnote{Period-of-day is associated with daylight, thus it is represented as ``day'' or ``night''.}, and {\em relevant points-of-interest data}. Importantly, we also present our \textit{process} for creating the above dataset from streaming traffic reports and heterogeneous contextual data (weather, points-of-interests, etc.), so that the community can validate it, and with the belief that this process can itself serve as a model for dataset creation. 
We performed a variety of data analysis and profiling based on US-Accidents dataset to derive a wide-range of insights. Our analyses demonstrated that about $40\%$ of accidents took place on or near high-speed roadways (highways, interstates, etc.) and about $32\%$ on or near local roads (streets, avenues, etc.). We also derived various insights with respect to the correlation of accidents with time, points-of-interest, and weather conditions. 

Using US-Accidents, and through extensive experiments across several large cities, we compared our proposal against several neural-network-based and traditional machine learning models (such as logistic regression and gradient boosting classifier). Our analysis and results show the superiority of our model in terms of improvement of $f1\text{-}score$ for the case of positive examples (i.e., cases which labeled as accident), by about $16\%$ in comparison to the best traditional model, and about $7\%$ in comparison to the best neural-network-based model. When considering both positive and negative cases (negative cases are labeled as non-accident, which are the majority), our proposal achieves comparable results when compared to the best baselines. Nevertheless, we note that positive cases are far more important, regarding their rare nature, and importance to be properly predicted. Further, we conducted thorough analyses to assess the ability of different categories of attributes for real-time traffic accident prediction using multiple testing scenarios. Our findings indicate the importance of time, points-of-interest, and traffic data for this task. 

The main contributions of this paper are therefore as follows. 
\begin{itemize}[leftmargin=*]
    \item A new methodology for heterogeneous data collection, cleansing, and augmentation to prepare a unique, large-scale dataset of traffic accidents. This dataset has been collected for the contiguous United States over three years, and contains $2.25$ million traffic accidents. The dataset is publicly available for the research community at \url{https://smoosavi.org/datasets/us\_accidents}. 
    \item A variety of insights gleaned through analyses of accident hot-spot locations, time, weather and points-of-interest correlations with the accident data. These insights may directly be utilized for applications such as urban planning, exploring flaws in transportation infrastructure design, traffic control and prediction, and personalized insurance.  
    \item A new deep-neural-network-based solution for traffic accident prediction using heterogeneous sparse data. To the best of our knowledge, this is the first work which uses information from {\em traffic flow}, fused with other available sources of contextual data such as ``weather'' and ``points-of-interest'', to perform accident prediction. Furthermore, our methodology predicts future accidents at the fine-grained time interval of 15 minutes. 
\end{itemize}

For the rest of this paper, we first provide preliminaries in Section~\ref{sec:prob}. The overview of related work is discussed in Section~\ref{sec:rel}. Section~\ref{sec:dataset} describes the dataset construction process and the resulting dataset. The accident prediction framework is presented in Section~\ref{sec:method}, followed by experiments and results in Section~\ref{sec:results}. Finally, Section~\ref{sec:conclusion} concludes the paper. 
\section{Related Work}
\label{sec:rel}
Accident analysis and prediction has been the topic of many research during the past few decades, where we study three categories of these work as follows. 

\vspace{5pt}
\noindent \textbf{Analysis of Environmental Stimuli on Accidents.} This category of work investigates the impact of environmental stimuli (e.g., weather, traffic flow, and properties of road-network) on possibility or severity of traffic accidents. Studying the impact of weather factors (e.g., precipitation) on road accidents \cite{eisenberg2004mixed,jaroszweski2014influence,tamerius2016precipitation, theofilatos2017incorporating}; applying data mining techniques to extract association rules to perform causality analysis \cite{kumar2015data,abellan2013analysis}; and statistical analysis of unobserved heterogeneity to explore the impact of unavailable variables (e.g., missing data) on severity of traffic accidents \cite{mannering2016unobserved} are some examples of this category. These studies usually provide significant insights, however, may not be directly utilized for real-time prediction and planning. 

\vspace{5pt}
\noindent \textbf{Accident Frequency Prediction.} Prediction of the expected number of traffic accidents for a specific road-segment or geographical region is the target of this group of studies \cite{chang2005data}. Early work in this area by Chang et al. \cite{chang2005analysis} used information such as road geometry, annual average daily traffic (AADT), and weather data to predict the frequency of accidents for a highway using a neural network model. Caliendo et al. \cite{caliendo2007crash} used a set of road-related attributes such as length, curvature, AADT, sight distance, and presence of junction to predict frequency of accidents. The usage of satellite imagery to predict the frequency of accidents by a convolutional neural network model using large scale accident and imagery data was proposed by Najjar et al. \cite{najjar2017combining}. Further, Ren et al. \cite{ren2018deep} recently used a Long Short Term Memory (LSTM) model to predict the frequency of accidents, given the history of past 100 hours, for grid cells of size $1km \times 1km$. Similarly, Chen et al. \cite{chen2018sdcae} proposed to use a stack denoising convolutional autoencoder model to predict frequency of accidents for grid cells using traffic flow (collected using plate recognition systems), past traffic accidents, and time data. Yuan et al. \cite{yuan2018hetero} proposed hetero-ConvLSTM to predict frequency of traffic accidents using several sources of environmental data such as traffic volume, road condition, rainfall, temperature, and satellite images. They evaluated their model using a large-scale data of traffic accidents from state of Iowa, performed predictions for grid cells of size $5km \times 5km$, and showed the importance of capturing spatial heterogeneity and temporal trends to better predict traffic accidents \cite{yuan2018hetero}. 
Studies in this category usually make use of many pieces of information that may not be available in real-time applications. 

\vspace{5pt}
\noindent \textbf{Accident Risk Prediction.} This category of work is very much similar to the previous one, unless prediction here is defined as a binary classification task which better fits real-time applications \cite{wenqi2017model,yuan2017predicting}. Using data for a single segment of I-64 in Virginia (US), Lin et al. \cite{lin2015novel} leveraged a decision tree model to separate pre-crash records from normal ones, using information such as weather, visibility, traffic volume, speed, and occupancy information. However, their limited size of data might weaken their solution or findings. In another study Chen et al. \cite{chen2016learning} used human mobility data in terms of 1.6 million GPS records and a set of 300,000 accident records in Tokyo (Japan) to predict the possibility of accident occurrence on grid cells of size $500 m \times 500 m$ in an hourly basis. They leveraged a stack denoising autoencoder model to extract latent features from human mobility, and then used a logistic regression model to predict accidents. Finally, Yuan et al. \cite{yuan2017predicting} used a heterogeneous set of urban data such as road characteristics (AADT, speed limit, etc.), radar-based rainfall data, temperature data, and demographic data to predict probability of accident for each road-segment in state of Iowa. They leveraged eigen-analysis to capture and represent spatial heterogeneity. Their analyses and results suggest the importance of time, human factors, weather data, and road-network characteristics for this task. 

Our proposal belongs to the last category as we seek to perform accident risk prediction. Further, our solution is more suitable for real-time applications as we provide prediction for much shorter time interval (i.e., 15 minutes) in comparison to literature. Besides, our usage of real-time traffic events and points-of-interest, to the best of our knowledge, is not discussed before. Lastly, the type of input data which we use for prediction is rather easy to collect and available to public, in contrast to those work which used extensive set of data for modeling and prediction. 
\section{Preliminaries and Problem Statement}
\label{sec:prob}

\begin{definition}[Traffic Event]
    We define a traffic event $e$ by $e = \langle lat, lng, time, type, desc \rangle$, where $lat$ and $lng$ represent the GPS coordinates, $type$ is a categorical classification of the event, and $desc$ provides a natural language description of the event. A traffic event is one of the following types: {\em accident}, {\em broken-vehicle}, {\em congestion}, {\em construction}, {\em event}, {\em lane-blocked}, and {\em flow-incident}. Table~\ref{tab:traffic_events} describes these events. 
\end{definition}

\begin{table}[ht]
    \vspace{-5pt}
    \small
    \setlength\tabcolsep{1pt}
    \centering
    \caption{\small Definition of Traffic Events.}\vspace{-10pt}
    \begin{tabular}{| c | c|}
        \rowcolor{Gray}
        \hline
        \textbf{Type} &  \textbf{Description}\\
        \hline
        Accident & A collision event which may involve one or more vehicles. \\ 
        \hline
        Broken-vehicle & \begin{tabular}{@{}c@{}} Refers to the situation when there is one \\ (or more) disabled vehicle(s) in a road. \end{tabular}\\ 
        \hline
        Congestion & \begin{tabular}{@{}c@{}} Refers to the situation when the speed of traffic \\  is slower than the expected speed or speed-limit. \end{tabular}\\ 
        \hline
        Construction & Refers to maintenance project on a road. \\ 
        \hline
        Event & \begin{tabular}{@{}c@{}} Situations such as {\em sports event}, {\em demonstrations}, or \\ {\em concerts}, that could potentially impact traffic flow. \end{tabular}\\ 
        \hline
        Lane-blocked & \begin{tabular}{@{}c@{}} Refers to the cases when we have blocked lane(s) \\ due to traffic or weather condition. \end{tabular}  \\ 
        \hline
        Flow-incident & \begin{tabular}{@{}c@{}} Refers to all other types of traffic events. \\ Examples are {\em broken traffic light} and {\em animal in the road}. \end{tabular}\\ 
        \hline
    \end{tabular}
    \label{tab:traffic_events}
    \vspace{-5pt}
\end{table}

\begin{definition}[Weather Observation Record]
    A weather observation $w$ is defined by $w = \langle lat, lng, time, temperature,$ $humidity,$ $pressure, visibility, wind\text{-}speed, precip,$ $rain, snow, fog,$ $hail \rangle$. Here $lat$ and $lng$ represent the GPS coordinates of the weather station which reported $w$; {\em precip} is the precipitation amount (if any); and rain, snow, fog, and hail are binary indicators of these events. 
\end{definition}

\begin{definition}[Point-of-Interest]
    A point-of-interest $p$ is defined by $p = \langle lat, lng, type \rangle$. Here, $lat$ and $lng$ show the GPS latitude and longitude coordinates, and available types for $p$ are described in Table~\ref{tab:poi_types}. Note that several of definitions in this table are adopted from \url{https://wiki.openstreetmap.org}. 
\end{definition}

\begin{table}[ht]
    \vspace{-5pt}
    \small
    \setlength\tabcolsep{1pt}
    \centering
    \caption{\small Definition of Point-Of-Interest (POI) annotation tags based on Open Street Map (OSM).}\vspace{-10pt}
    \begin{tabular}{| c | c|}
        \rowcolor{Gray}
        \hline
        \textbf{Type} &  \textbf{Description}\\
        \hline
        Amenity & \begin{tabular}{@{}c@{}} Refers to particular places such as restaurant,\\ library, college, bar, etc.\end{tabular} \\ 
        \hline
        Bump & Refers to speed bump or hump to reduce the speed. \\ 
        \hline
        Crossing & \begin{tabular}{@{}c@{}} Refers to any crossing across roads for \\ pedestrians, cyclists, etc.\end{tabular} \\ 
        \hline
        Give-way & A sign on road which shows priority of passing. \\ 
        \hline
        Junction & Refers to any highway ramp, exit, or entrance. \\ 
        \hline
        No-exit & \begin{tabular}{@{}c@{}} Indicates there is no possibility to travel further\\ by any transport mode along a formal path or route.\end{tabular}  \\ 
        \hline
        Railway & Indicates the presence of railways. \\ 
        \hline
        Roundabout & Refers to a circular road junction.\\  
        \hline
        Station & Refers to public transportation station (bus, metro, etc.). \\ 
        \hline
        Stop & Refers to stop sign. \\ 
        \hline
        Traffic Calming & Refers to any means for slowing down traffic speed. \\ 
        \hline
        Traffic Signal & Refers to traffic signal on intersections. \\ 
        \hline
        Turning Loop & \begin{tabular}{@{}c@{}} Indicates a widened area of a highway with\\a non-traversable island for turning around. \end{tabular}  \\ 
        \hline
    \end{tabular}
    \label{tab:poi_types}
    \vspace{-5pt}
\end{table}

\begin{definition}[Geographical Region]
    We define a geographical region $r$ as a square of size $l\times l$ over the map of a city. The choice of $l$ is related to application domain, and in this work we set $l = 5km$. 
\end{definition}

Given the preliminaries, we formulate the problem as follows: 

\vspace{3pt}
\noindent\textbf{Given:}
\begin{itemize}
    \item [--] A spatial grid $\mathrm{R} = \{r_1, r_2, \dots, r_n\}$, where each $r \in \mathrm{R}$ is a geographical region of size $5km \times 5km$. 
    \item [--] A set of fixed-length time intervals $\mathrm{T} = \{t_1, t_2, \dots, t_m\}$, where we set $|t| = 15$ {\it minutes}, for $t \in \mathrm{T}$.
    \item [--] A database of traffic events $E_r = \{e_1, e_2, \dots\}$ for each geographical region $r \in \mathrm{R}$. 
    \item [--] A database of weather observation records $W_r = \{w_1, w_2, \dots\}$ for each geographical region $r \in \mathrm{R}$. 
    \item [--] A database of points of interest $P_r = \{p_1, p_2, \dots\}$ for each geographical region $r \in \mathrm{R}$. 
\end{itemize}

\noindent\textbf{Create:}\vspace{-5pt}
\begin{itemize}
    \item [--] A representation $F_{rt}$ for a region $r \in \mathrm{R}$ during a time interval $t \in \mathrm{T}$, using $E_r$, $W_r$, and $P_r$.
    \item [--] A binary label $L_{rt}$ for $F_{rt}$, where 1 indicates at least one traffic accident happened during $t$ in region $r$, and 0 otherwise. 
\end{itemize}

\noindent\textbf{Find:}\vspace{-5pt}
\begin{itemize}
    \item [--] A model $\mathcal{M}$ to predict $L_{rt}$ using $\langle F_{rt_{i-8}}, F_{rt_{i-7}}, \dots, F_{rt_{i-1}} \rangle$, which means predicting the label of current time interval using observations from the last 8 time intervals to   
\end{itemize}

\noindent\textbf{Objective:}\vspace{-5pt}
\begin{itemize}
    \item [--] Minimize the prediction error.  
\end{itemize}
\section{Accident Dataset}
\label{sec:dataset}
\begin{figure}[ht]
    \vspace{-5pt}
    \centering
    \includegraphics[scale=0.45]{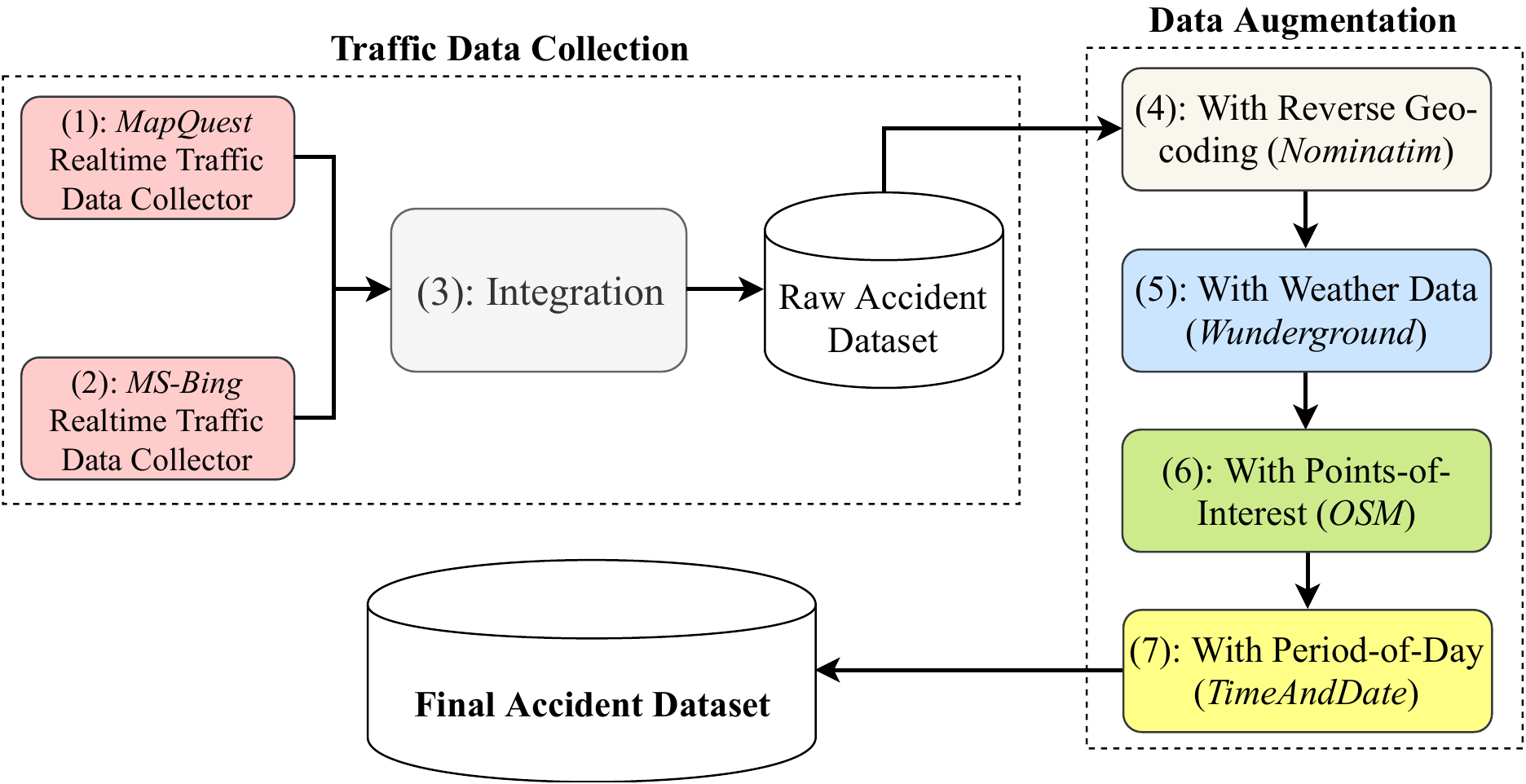}
    \caption{Process of Creating Traffic Accident Dataset}
    \vspace{-5pt}
    \label{fig:data_process}
    \vspace{-8pt}
\end{figure}

This section describes the process of constructing a country-wide traffic accident dataset, which we named {\em US-Accidents}. An overview of this process is shown in Figure~\ref{fig:data_process}. US-Accident contains $2.25$ million cases of traffic accidents that took place within the United States from February 2016 to March 2019. The following sub-sections provide a detailed description of each step of the data preparation process. The dataset is publicly available at \url{https://smoosavi.org/datasets/us\_accidents}.

\subsection{Traffic Data Collection}

\subsubsection{Realtime Traffic Data Collection}
We collected streaming traffic data using two real-time data providers, namely ``MapQuest Traffic'' \cite{mapquest} and ``Microsoft Bing Map Traffic'' \cite{bing}, whose APIs broadcast traffic events (accident, congestion, etc.) captured by a variety of entities - the US and state departments of transportation, law enforcement agencies, traffic cameras, and traffic sensors within the road-networks. We pulled data every 90 seconds from 6am to 11pm, and every 150 seconds from 11pm to 6am. In total, we collected $2.27$ million cases of traffic accidents between February 2016 and March 2019; $1.73$ million cases were pulled from MapQuest, and $0.54$ million cases from Bing. 

\subsubsection{Integration}
Integration of the data consisted of removing cases duplicated across the two sources and building a unified dataset. We considered two events as duplicates if their Haversine distance and their recorded times of occurrence were both below a heuristic threshold (set empirically at 250 meters and 10 minutes, respectively). We believe these settings to be conservative, but we settled on them in order to ensure a very low possibility of duplicates. Using these settings, we found about $24,600$ duplicated accident records, or about $1\%$ of all data. The final dataset after removing the duplicated cases comprised $2.25$ million accidents. 

\subsection{Data Augmentation}
\subsubsection{Augmenting with Reverse Geo-Coding}
Raw traffic accident records contained only GPS data. We employed the {\em Nominatim} tool \cite{nominatim} to perform reverse geocoding to translate GPS coordinates to addresses, each consisting of a {\em street number}, {\em street name}, {\em relative side (left/right)}, {\em city}, {\em county}, {\em state}, {\em country}, and {\em zip-code}. This process is same as {\em point-wise map-matching}. 

\subsubsection{Augmenting with Weather Data}
Weather information provides important context for traffic accidents. Thus, we employed {\em Weather Underground} API  \cite{wunderground} to obtain  weather information for each accident. Raw weather data was collected from 1,977 weather stations located in airports all around the United States. The raw data comes in the form of observation records, where each record consists of several attributes such as {\em temperature}, {\em humidity}, {\em wind-speed}, {\em pressure}, {\em precipitation} (in millimeters), and {\em condition}\footnote{Possible values are {\em clear}, {\em snow}, {\em rain}, {\em fog}, {\em hail}, and {\em thunderstorm}.}. For each weather station, we collected several data records per day, each of which was reported upon any significant change in any of the measured weather attributes. 

Each traffic event $e$ was augmented with weather data as follows. First the closest weather station $s$ was identified. Then, of the weather observation records which were reported from $s$, we looked for the weather observation record $w$ whose reported time was closest to the start time of $e$, and augmented it with weather data. In our integrated accident dataset, the average difference in report time for an accident record and its paired weather observation record was about $15$ minutes. 

\subsubsection{Augmenting with Points-Of-Interest}
Points-of-interest (POI) are locations annotated on a map as {\em amenities}, {\em traffic signals}, {\em crossings}, etc. These annotations are associated with {\em nodes} on a road-network. A node can be associated with a variety of POI types, however, in this work we only use 13 types as described in Table~\ref{tab:poi_types}. We obtained these annotations from Open Street Map (OSM) \cite{osm} for the United States, using its most recently released dataset (extracted on April 2019). The applicable POI annotations for a traffic accident $a$ are those which are located within a distance threshold $\tau$ from $a$. We determine this threshold by evaluating different values to find the value that is best able to associate a POI with an accident. Essentially, the objective is to find the best distance for which a POI annotation can be identified as {\it relevant} to an accident record. Therefore, we need a mechanism to measure the relevancy. To begin with, we note that the natural language descriptions of traffic accidents follow a set of regular expression patterns, and that a few of these patterns may be used to identify and use as an annotation for the location type (e.g., intersection or junction) of the accident. 

\vspace{6pt}
\noindent \textbf{Regular Expression Patterns.} Given the description of traffic accidents, we were able to identify 27 regular expression patterns; 16 of them were extracted based on MapQuest data, and 11 from Bing data. Among the MapQuest patterns, the following expression corresponds to {\em junctions} (see Table~\ref{tab:poi_types}): ``$\dots$ \textbf{on} $\dots$ \textbf{at exit} $\dots$'', and the following pattern mostly\footnote{Using 200 randomly sampled accidents cases which were manually checked on a map, about 78\% of matches using this pattern were actually occurred on intersections.} determines an {\em intersection}: ``$\dots$ \textbf{on} $\dots$ \textbf{at} $\dots$''. An intersection is associated with {\em crossing}, {\em stop}, or {\em traffic signal} (see Table~\ref{tab:poi_types}). Among Bing regular expression patterns, two of them identify junctions: ``\textbf{at} $\dots$ \textbf{exit} $\dots$'' and ``\textbf{ramp to} $\dots$''. Table~\ref{tab:accident_examples} shows several examples of accidents, where the regular expression pattern (in bold face) identifies the correct POI type\footnote{These cases were manually checked on a map to ensure the correctness of the annotation.}. 

\begin{table}[ht]
    \vspace{-4pt}
    \small
    \centering
    \setlength\tabcolsep{2pt}
    \caption{\small Examples of traffic accidents with their {\em annotation type} assigned using their natural language description by regular expression patterns.}\vspace{-5pt}
    \begin{tabular}{ c | c | c }
        \textbf{Source} & \textbf{Description} & \textbf{Type}\\
        \hline
        MapQuest & Serious accident \textbf{on} 4th Ave \textbf{at} McCullaugh Rd. & Intersection\\
        MapQuest & Accident \textbf{on} NE-370 Gruenther Rd \textbf{at} 216th St. & Intersection\\
        MapQuest & Accident \textbf{on} I-80 \textbf{at Exit} 4A Treasure Is. & Junction\\
        MapQuest & Accident \textbf{on} I-87 I-287 Southbound \textbf{at Exit} 9 I-287. & Junction\\
        Bing & \textbf{At} Porter Ave/\textbf{Exit} 9 - Accident. Left lane blocked. & Junction\\
        Bing & \textbf{At} IL-43/Harlem Ave/\textbf{Exit} 21B - Accident. & Junction\\
        Bing & \textbf{Ramp to} I-15/Ontario Fwy/Cherry Ave - Accident. & Junction\\
        Bing & \textbf{Ramp to} Q St - Accident. Right lane blocked. & Junction\\
    \end{tabular}
    \label{tab:accident_examples}
\end{table}

\begin{algorithm}[ht]
    \small
    \caption{Find Annotation Correlation}
    \begin{algorithmic}[1]
        \STATE Input: a dataset of traffic accidents $\mathcal{A}$, a database of points-of-interest $\mathcal{P}$, and a distance threshold $\tau$.
        \STATE Extract and create a set of regular expression patterns $RE$ to identify a specific POI $\nu$. 
        \STATE Create set $S_1$: for each traffic accident $a \in \mathcal{A}$, we add it to $S_1$ if its natural language description $a.desc$ can be matched with at least one regular expression in set $RE$. 
        \STATE Create set $S_2$: for each traffic accident $a \in \mathcal{A}$, we add it to $S_2$ if there is at least one POI $p \in \mathbf{P}$ of type $\nu$, where $\textit{haversine\_distance }(a, p) \leq \tau$.
        \STATE Output: Return $\textit{Jaccard }(S_1, S_2)$. 
    \end{algorithmic}
    \label{alg:find_tau}
\end{algorithm}

\begin{figure*}[ht]
    \small
    \centering
    \hspace{-25pt}
    \begin{subfigure}[b]{0.3\textwidth}
            \includegraphics[width=\linewidth]{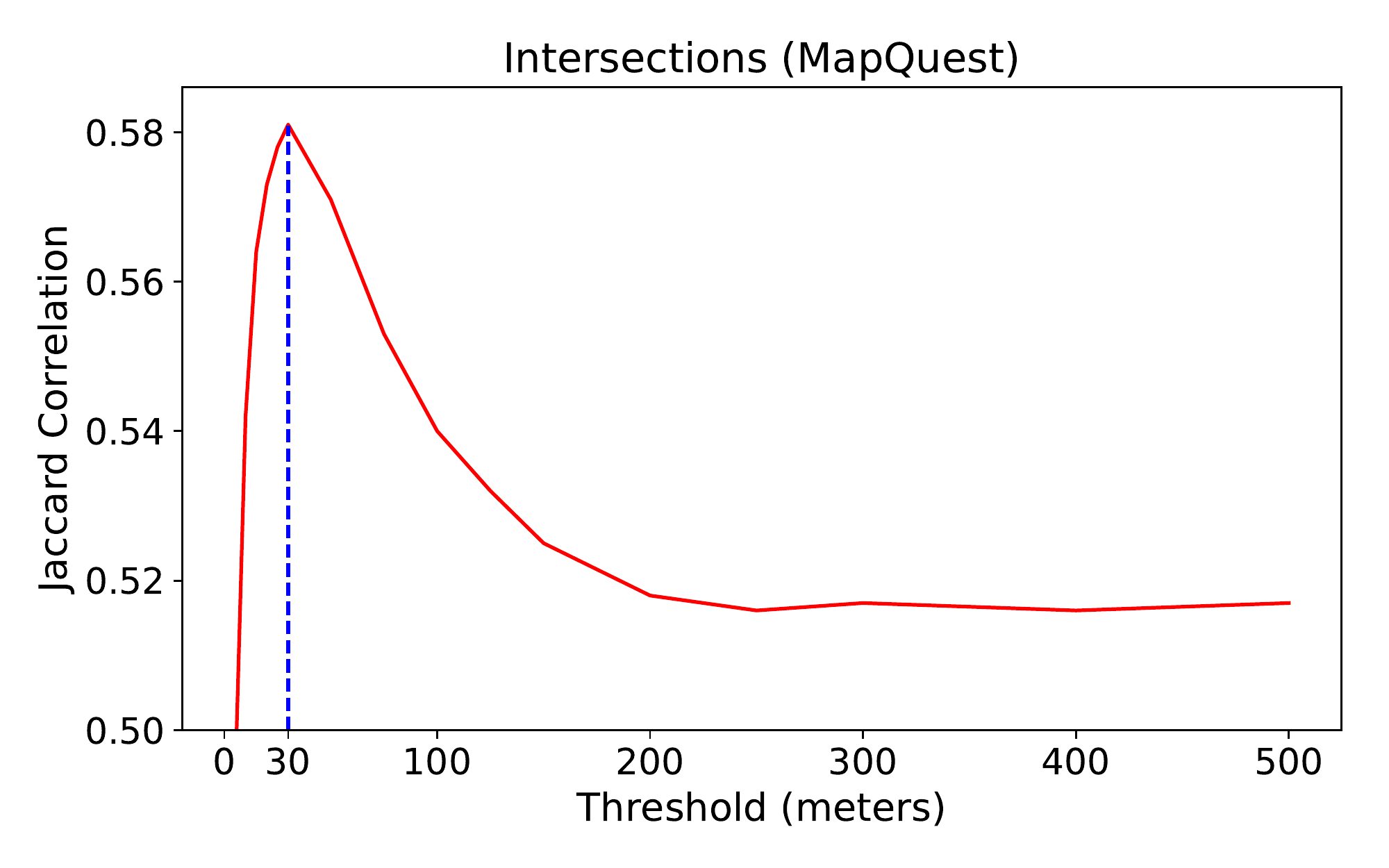}
            \caption{\small Using MapQuest for Intersection}
            \label{fig:mq_intersec}
    \end{subfigure}\hspace{-5pt}
    \begin{subfigure}[b]{0.3\textwidth}
            \includegraphics[width=\linewidth]{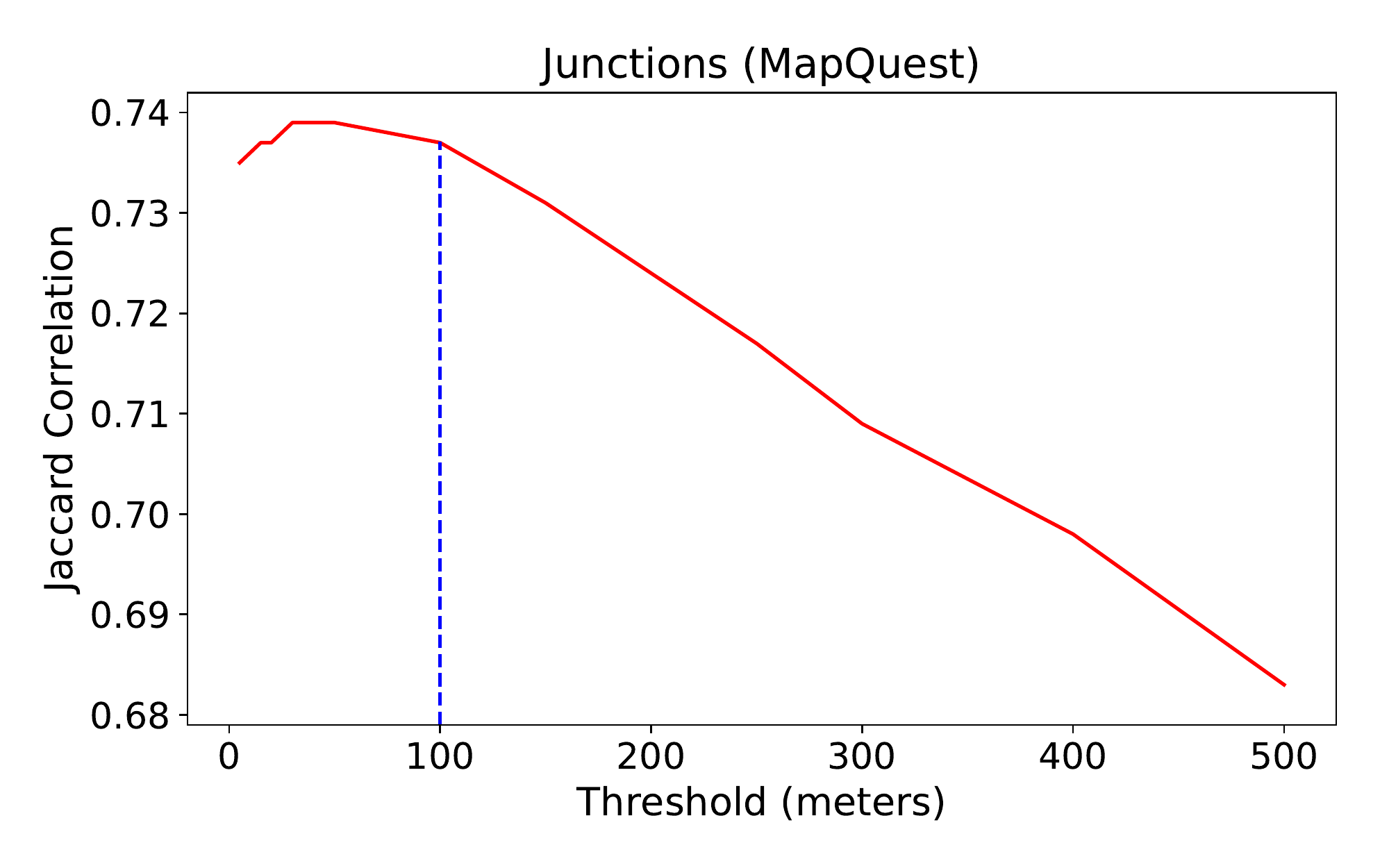}
            \caption{\small Using MapQuest for Junction}
            \label{fig:mq_junction}
    \end{subfigure}\hspace{-5pt}
    \begin{subfigure}[b]{0.3\textwidth}
            \includegraphics[width=\linewidth]{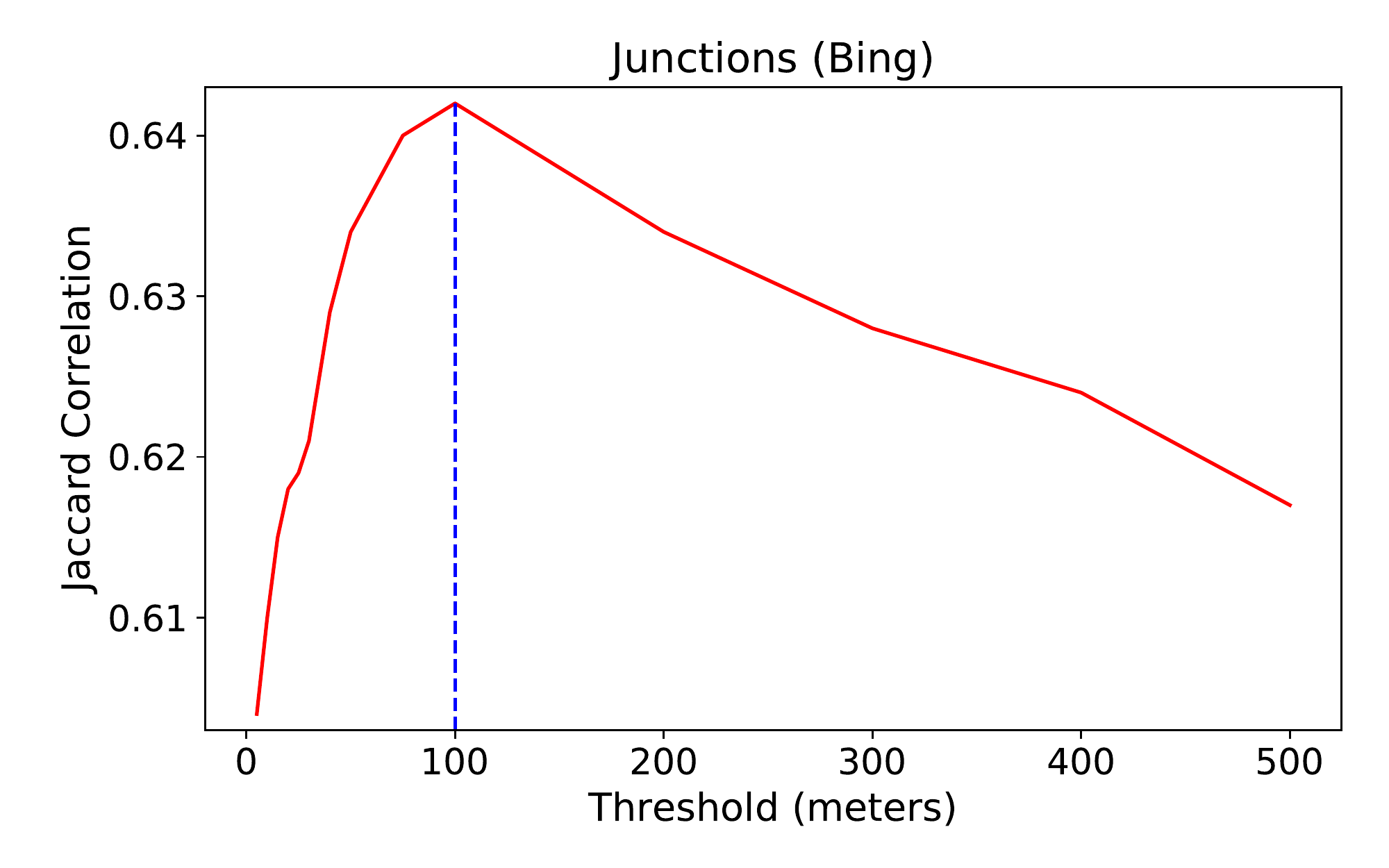}
            \caption{\small Using Bing for Junction}
            \label{fig:bg_junction}
    \end{subfigure}
    \hspace{-22pt}
    \vspace{-10pt}
    \caption{\small Correlation study between regular-expression and OSM-based extracted annotations to find the best distance threshold values.}
    \label{fig:text_tag_correlation}
    \vspace{-5pt}
\end{figure*}

\vspace{6pt}
The essential idea is to find a threshold value that maximizes the correlation between annotations from POI and annotations derived using regular expression patterns. Thus, for a set of accident records, we annotate their location based on both methods, regular expression patterns as well as OSM-based POI annotations (using a specific distance threshold). Then, we measure the correlation between the annotations derived from these methods to find which threshold value provides the highest correlation (i.e., the best choice). Note that we employ the regular expression patterns as \textit{pseudo} ground truth labels, to evaluate OSM-based POI annotations using different threshold values. We propose Algorithm~\ref{alg:find_tau} to find the best distance threshold. We use a sample of $100,000$ accidents as set $\mathcal{A}$ (step 1). For step 2, we consider either ``intersection'' or ``junction'', and use the set of relevant regular expressions (see Table~\ref{tab:accident_examples}) in terms of $RE$. Next we create set $S_1$ by annotating each traffic accident $a \in \mathcal{A}$ using the regular expression patterns in $RE$ (step 3). Then we annotate each traffic accident $a \in \mathcal{A}$ based on points-of-interests in $\mathcal{P}$, using the distance threshold $\tau$ to create $S_2$ (step 4). Finally, we calculate the Jaccard similarity score using Equation~\ref{eq:jaccard} (step 5):
\begin{equation}
    \label{eq:jaccard}
    \textit{Jaccard }(S_1, S_2) = \frac{\big|\, S_1 \cap S_2\, \big|}{\big|\, S_1 \cup S_2\, \big|}
\end{equation}
We examined the following candidate set to find the optimal threshold value (all values in meters): $\{5, 10, 15, 20, 25, 30, 40, 50, 75, 100, 125,$ $150, 200, 250, 300, 400, 500\}$. We separately studied samples from Bing and MapQuest, and employed corresponding regular expression patterns for ``intersection'' and ``jucntion''. Figure~\ref{fig:text_tag_correlation} shows the results for each data source and each annotation type. From Figure~\ref{fig:mq_intersec}, we see that the maximum correlation for intersections is obtained for a threshold value of 30 meters. Figures~\ref{fig:mq_junction} and \ref{fig:bg_junction} show that 100 meters is an appropriate distance threshold for annotating a junction. 

\begin{figure*}[h]
    \minipage{0.3\textwidth}
        \centering
        \includegraphics[width=\linewidth]{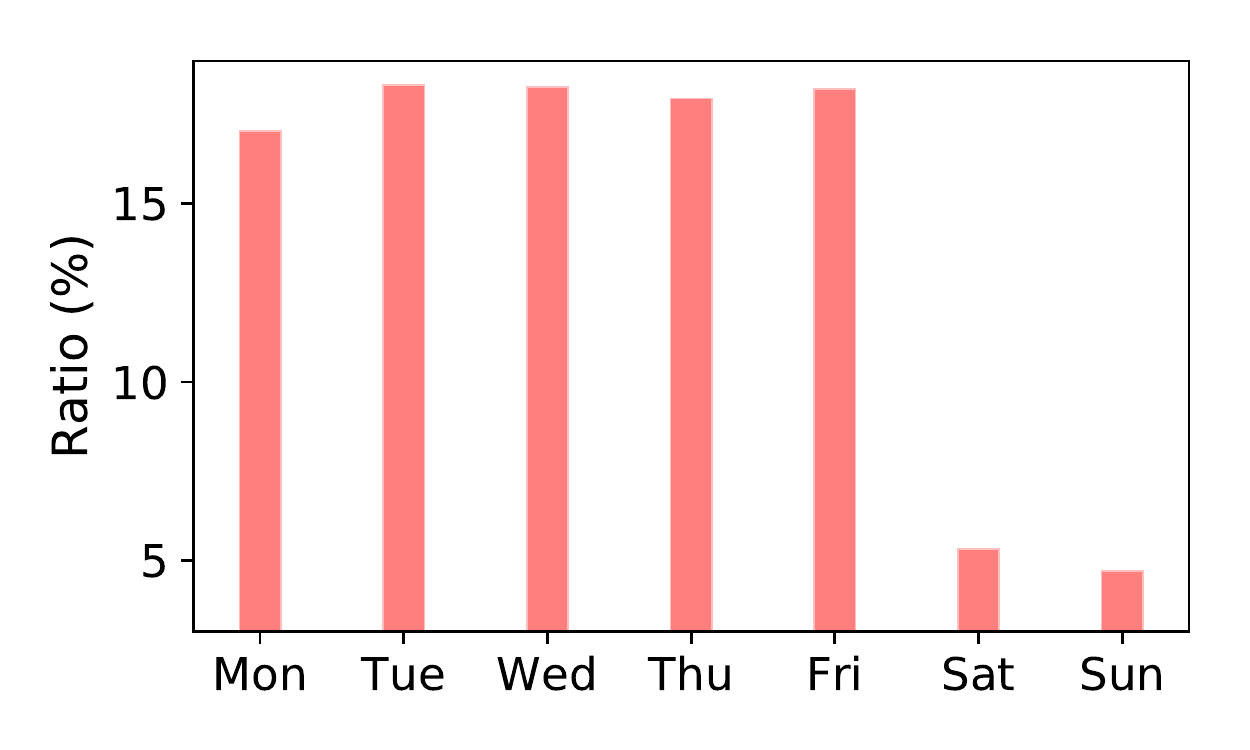}
        (a) Day of Week
    \endminipage\hspace{5pt}
    \minipage{0.3\textwidth}
        \centering
        \includegraphics[width=\linewidth]{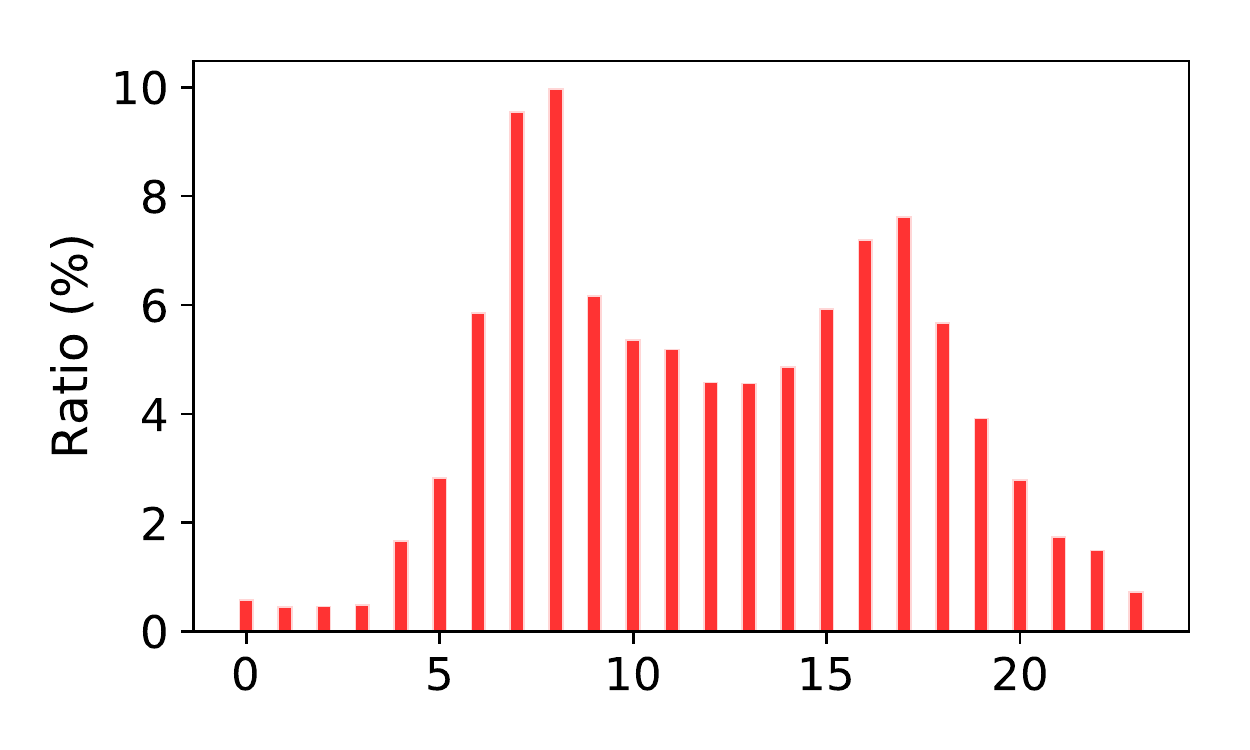}
        (b) Hour of Day (weekdays)
    \endminipage\hspace{5pt}
    \minipage{0.3\textwidth}
        \centering
        \includegraphics[width=\linewidth]{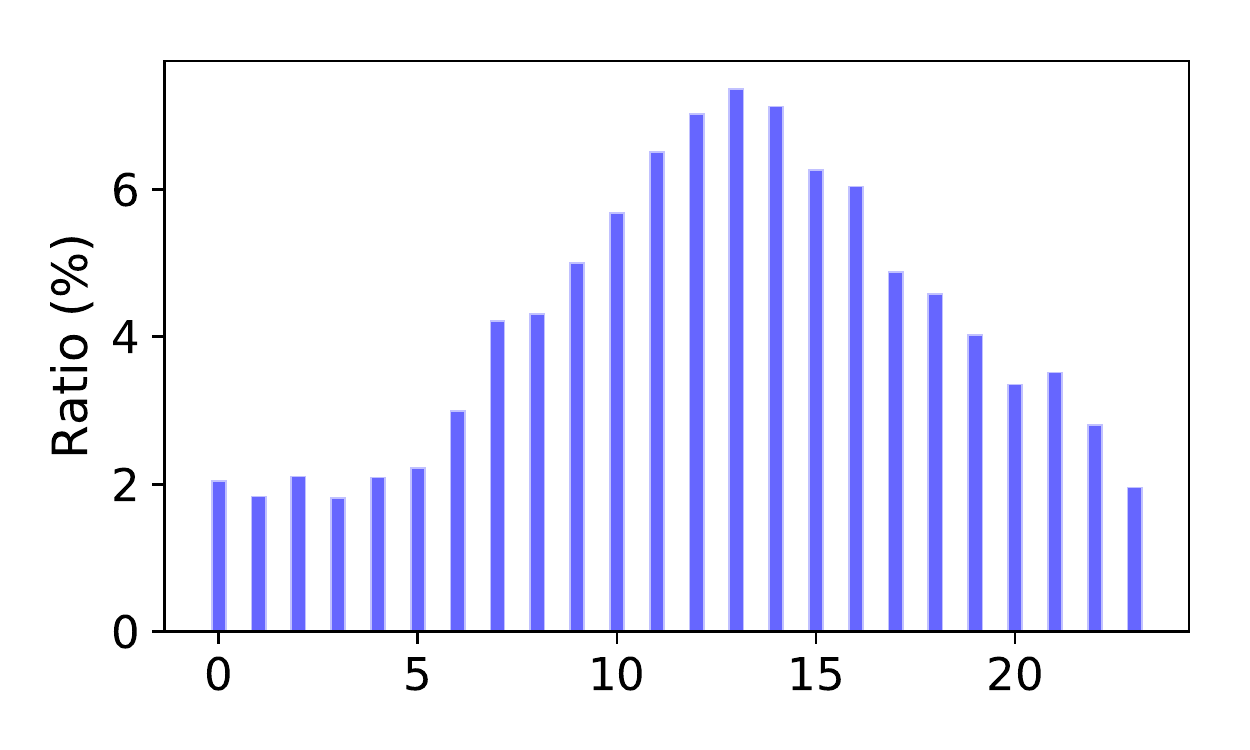}
        (c) Hour of Day (weekends)
    \endminipage\hspace{5pt}
    \minipage{0.3\textwidth}
        \centering
        \includegraphics[width=\linewidth]{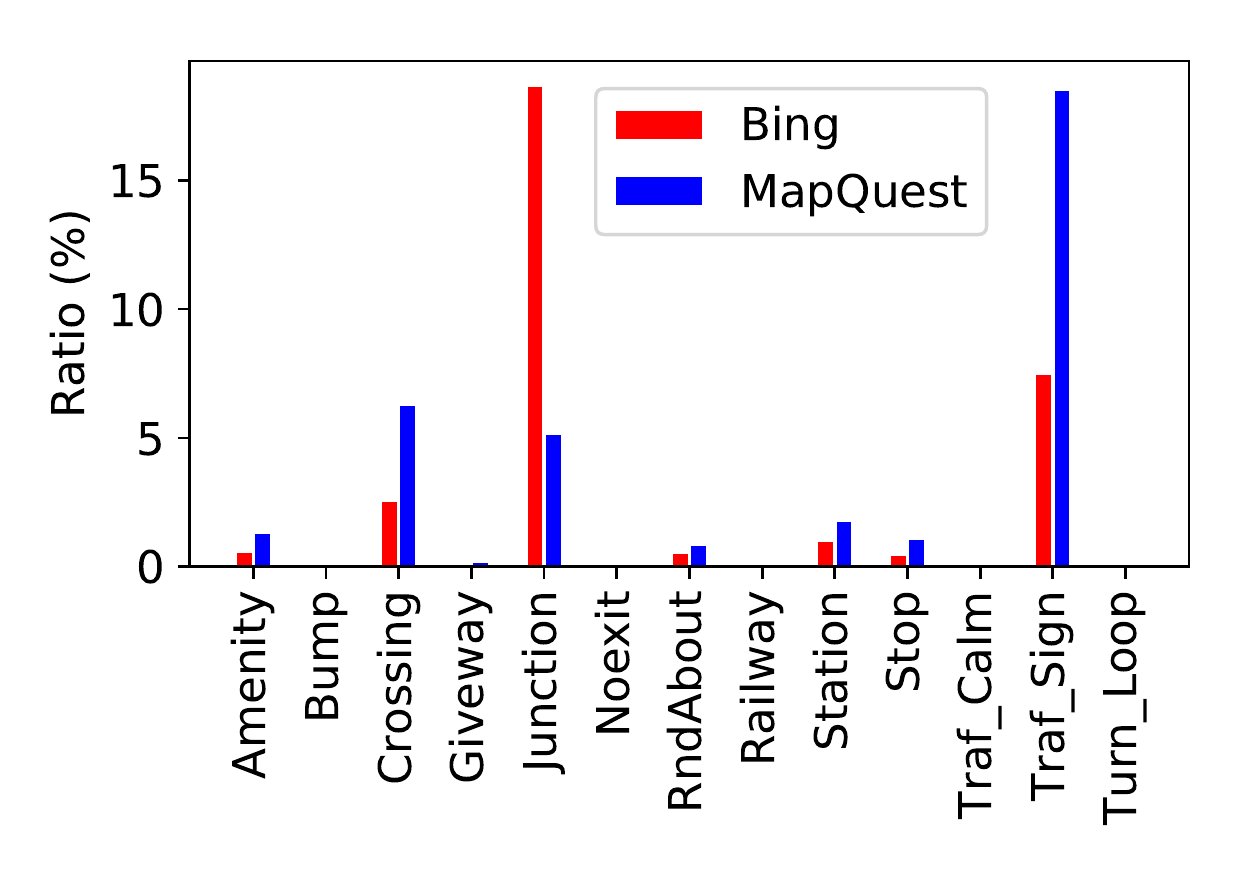}
        (d) points-of-interest Distribution
    \endminipage\hspace{5pt}
    \minipage{0.3\textwidth}
        \centering
        \includegraphics[width=\linewidth]{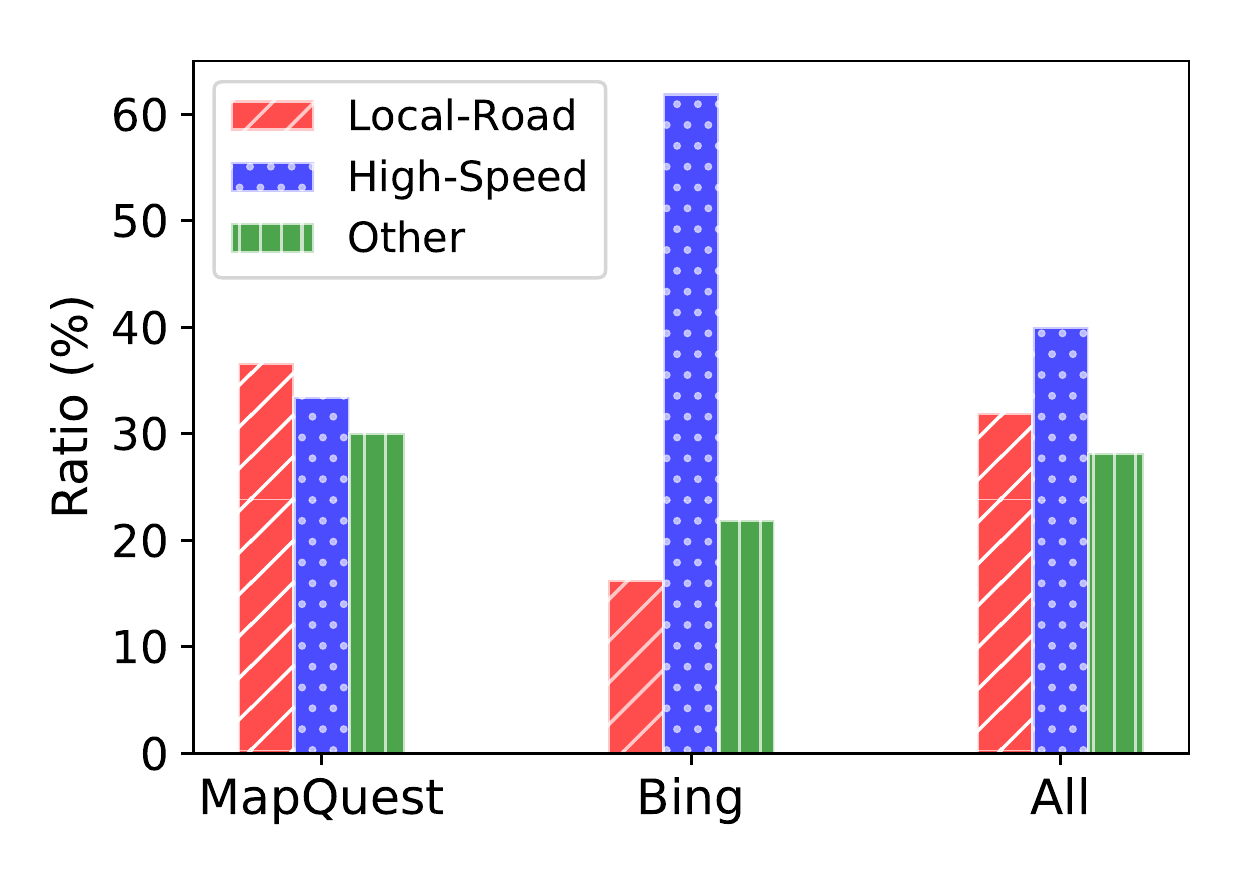}
        (e) Road-type Distribution
    \endminipage\hspace{5pt}
    \minipage{0.3\textwidth}
        \centering
        \includegraphics[width=\linewidth]{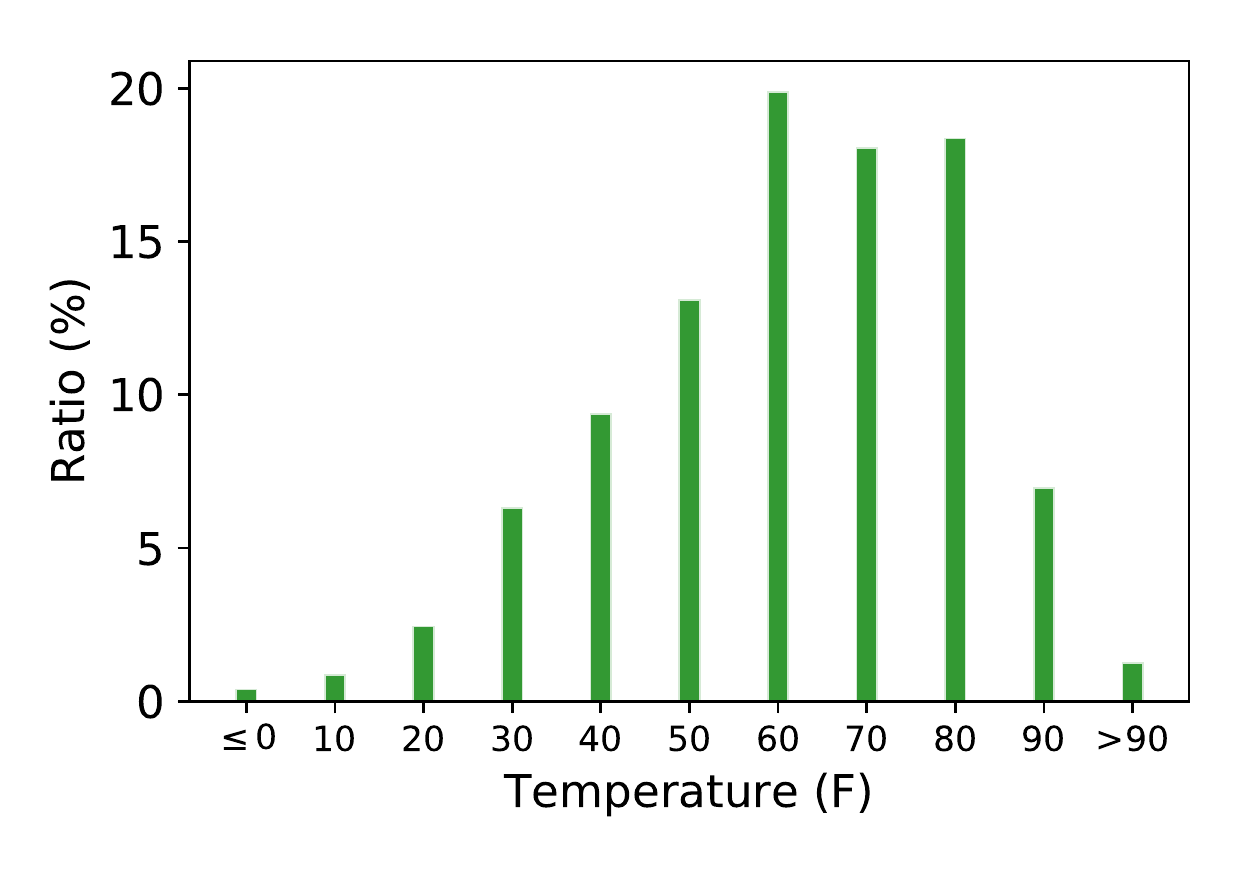}
        (f) Temperature Distribution
    \endminipage\hspace{5pt}
    \vspace{-8pt}
    \minipage{\textwidth}
        \centering
        \caption{Characteristics of US-Accidents dataset, in terms of time analysis (a)--(c), points-of-interest-based augmentation distribution analysis (d), map-matching-based road type coverage analysis (e), and temperature analysis (f).}
        \label{fig:all_profilings}
    \endminipage\hfill
\end{figure*}

Thresholds for the other available annotations in Table~\ref{tab:poi_types} are derived from the thresholds for junction and intersection as described below:
\begin{itemize}[leftmargin=*]
    \item \textbf{Junction-based threshold.} Given the definition of a junction (i.e., a highway ramp, exit, or entrance), we used the same threshold (100 meters) for the following types: amenity and no-exit. 
    \item \textbf{Intersection-based threshold.} Given the definition of an intersection, we used the same threshold (30 meters) for the following annotation types: bump, crossing, give-way, railway, roundabout, station, stop, traffic calming, traffic signal, and turning loop. 
\end{itemize}
Using these thresholds, we augmented each accident record with points-of-interest. In summary, $27.5 \%$ of accident records were augmented with at least one of the available POI types in Table~\ref{tab:poi_types}. Further discussion on annotation results are presented in Section~\ref{subsec:us_accidents}. 

\subsubsection{Augmenting with Period-of-Day}
Given the start time of an accident record, we used ``TimeAndDate'' API \cite{timeanddate} to label it as {\em day} or {\em night}. We assign this label based on four different daylight systems, namely {\em Sunrise/Sunset}, {\em Civil Twilight}, {\em Nautical Twilight}, and {\em Astronomical Twilight}. Note that these systems are defined based on the position of the sun with respect to the horizon, and each provide a different definition for period-of-day\footnote{See \url{https://en.wikipedia.org/wiki/Twilight} for more details.}. 

\subsection{US-Accidents Dataset}
\label{subsec:us_accidents}
Using the process described above, we created a countrywide dataset of traffic accidents, which we name {\em US-Accidents}. US-Accident contains about $2.25$ million cases of traffic accidents that took place within the contiguous United States from February 2016 to March 2019. Table~\ref{tab:data_facts} shows the important details of US-Accidents. Also, Figure~\ref{fig:all_profilings} provides more details on characteristics of the dataset. Figure~\ref{fig:all_profilings}-(a) shows that significantly more accidents were observed during the weekdays than weekends. Based on parts (b) and (c) of Figure~\ref{fig:all_profilings}, it can be observed that the hourly distribution during weekdays has two peaks (8am and 5pm), while the weekend distribution shows a single peak (1pm). Figure~\ref{fig:all_profilings}-(d) demonstrates that most of the accidents took place near junctions or intersections (crossing, traffic signal, and stop). MapQuest tends to report more accidents near intersections, while Bing reported more cases near junctions. This shows the complementary behavior of these APIs, and hence the comprehensiveness of our dataset. 
Figure~\ref{fig:all_profilings}-(e) describes distribution of road types, extracted from the map-matching results (i.e., street names). We used street names to identify type of the road. Here we note that about $32\%$ of accidents happened on or near local roads (e.g., streets, avenues, and boulevards), and about $40\%$ took place on or near high-speed roads (e.g., highways, interstates, and state roads). We also note that Bing reported more cases on high-speed roads. Finally, the period-of-day data shows that about $73\%$ of accidents happened after sunrise (or during the day). 

\begin{table}[ht]
    \small
    \setlength\tabcolsep{1pt}
    \centering
    \caption{\small US-Accidents: details as of March 2019.} \vspace{-5pt}
    \begin{tabular}{|>{\columncolor[gray]{0.9}}c|c|}
        \hline
        Total Attributes &  45\\
        \hline
        Traffic Attributes (10) &  \begin{tabular}{@{}c@{}} id, source, TMC \cite{tmc}, severity, start\_time, end\_time, \\ start\_point, end\_point, distance, and description \end{tabular} \\
        \hline
        Address Attributes (8) & \begin{tabular}{@{}c@{}} number, street, side (left/right), city, \\ county, state, zip-code, country  \end{tabular}\\
        \hline
        Weather Attributes (10) & \begin{tabular}{@{}c@{}} time, temperature, wind\_chill, humidity,\\ pressure, visibility, wind\_direction, wind\_speed,\\precipitation, and condition (e.g., rain, snow, etc.) \end{tabular}\\
        \hline
        POI Attributes (13) & All cases in Table~\ref{tab:poi_types}\\
        \hline
        Period-of-Day (4) & \begin{tabular}{@{}c@{}} Sunrise/Sunset, Civil Twilight, \\Nautical Twilight, and Astronomical Twilight\end{tabular}\\
        \hline
        \hline
        Total  Accidents &  2,243,939\\
        \hline
        \# MapQuest Accidents & 1,702,565 (75.9\%) \\
        \hline
        \# Bing Accidents & 516,762 (23\%)\\
        \hline
        \# Reported by Both & 24,612 (1.1\%) \\
        \hline
        Top States & \begin{tabular}{@{}c@{}}  California (485K), Texas (238K), Florida (177K), \\ North Carolina (109K), New York (106K) \end{tabular}\\
        \hline
    \end{tabular}
    \label{tab:data_facts}
\end{table}


\section{Accident Prediction Model}
\label{sec:method}
In this section we describe our traffic accident prediction framework. We start with description of feature vector representation, and then 
present our proposal for real-time traffic accident prediction. 

\subsection{Feature Vector Representation}
\label{subsec:feature_vec}
Regarding the problem description in Section~\ref{sec:prob}, we create a feature vector representation for each geographical region $r$ of size $5km \times 5km$ during a time interval $t = 15\, minutes$. Such representation includes the following feature categories:
\begin{itemize}[leftmargin=*]
    \item \textbf{Traffic}: a vector of size 7 representing frequency of available traffic events (i.e., accident, broken-vehicle, congestion, construction, event, lane-blocked, and flow-incident) during the current 15 minutes interval. We obtain traffic events from \cite{moosavi2019short}. 
    \item \textbf{Time}: includes {\em weekday} (a binary value to show weekday or weekend), {\em hour-of-day} (a one-hot vector of size 5 to show belonging to a specific time interval as defined in \cite{moosavi2017characterizing})\footnote{These time intervals are [6am -- 10am], [10am -- 3pm], [3pm -- 7pm], [7pm -- 10pm], and [10pm -- 6am].}, and {\em daylight} (an attribute to show period-of-day: day or night). We obtain daylight data from \cite{timeanddate}. 
    \item \textbf{Weather}: a vector representing 10 weather attributes including temperature, pressure, humidity, visibility, wind-speed, precipitation amount; and four indicator flags for special events rain, snow, fog, and hail. We obtain weather data from \cite{wunderground}. 
    \item \textbf{POI}: a vector of size 13 to represent frequency of POIs within $r$, for amenity, speed bump, crossing, give-way sign, junction, no-exit sign, railway, roundabout, station, stop sign, traffic calming, traffic signal, and turning loop. We obtain POI data from \cite{osm}. 
    \item \textbf{Desc2Vec}: given a historical set of traffic events in region $r$, we use their natural language description, and by employing the GloVe pre-trained distributed word vectors \cite{pennington2014glove}, we create a description to vector (Desc2Vec) representation for $r$. Such representation is the average representation of words in description of all events which took place within $r$ during a particular time period. Size of this vector is $100$. The choice of GloVe among the existing models is because of its well-known applicability for generic applications and also reasonable dictionary size (i.e., 400K terms). We obtain traffic events from \cite{moosavi2019short}. 
\end{itemize}

\noindent In this way, we represent $r$ during time interval $t$ by 24 time-variant (i.e., traffic, time, and weather) and 113 time-invariant (i.e., POI and Desc2Vec) attributes. In order to predict the label of $r$ during $t$, we use a vector representing the last 8 time intervals (last two hours), including one instance of time-invariant attributes (113 features) and 8 instances of time-variant attributes ($8 \times 24$ features)\footnote{See Section~\ref{sec:prob} for formulation of prediction task.}. 

\subsection{Deep Accident Prediction (DAP) Model}
\label{subsec:dap}
\begin{figure}[ht]
    \centering
    \hspace{-17pt}
    \includegraphics[scale=0.5]{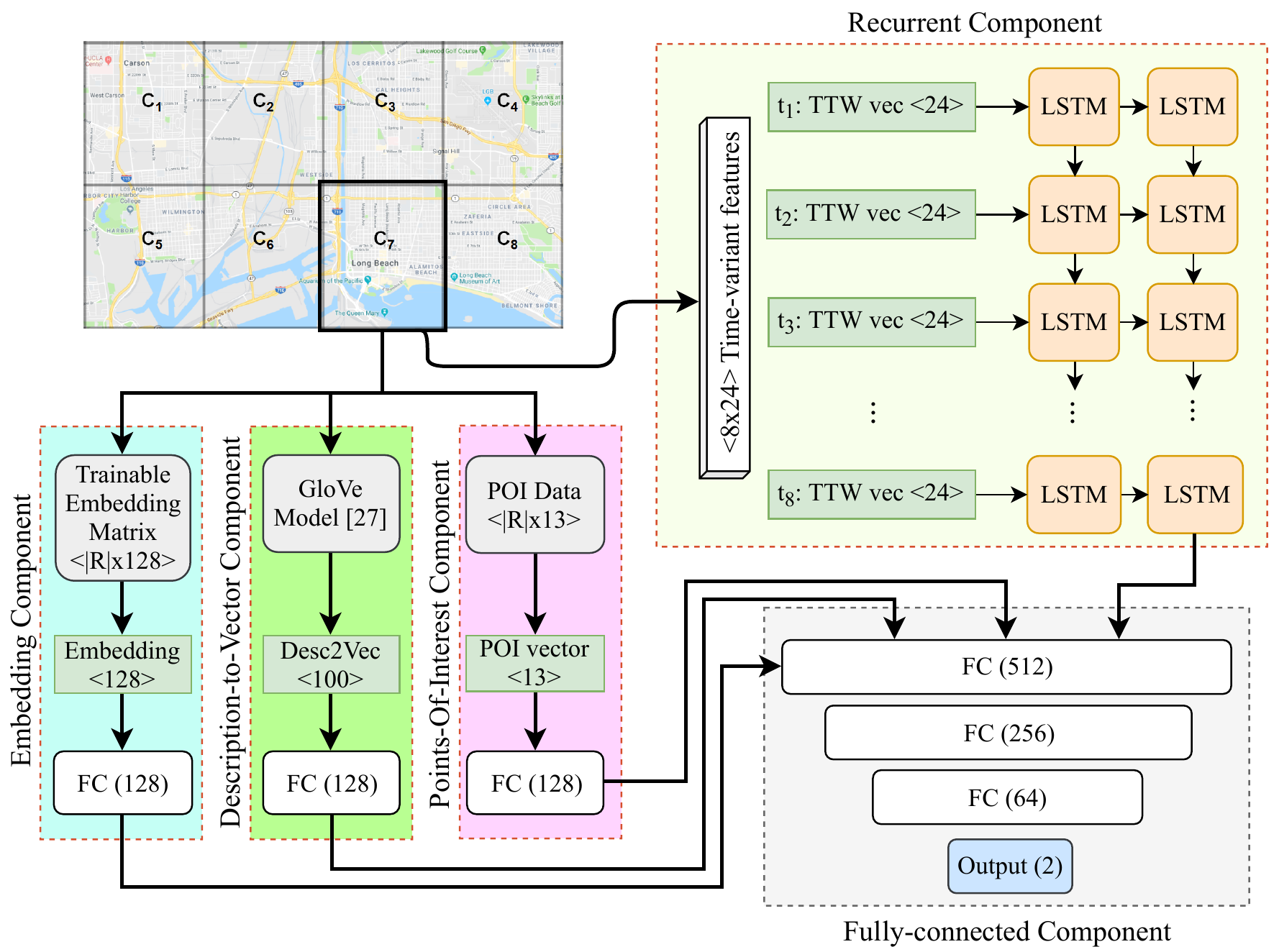}
    \caption{DAP: A Deep neural-network-based Accident Prediction model. Here, R is the set of all regions; each $C_i$ is a grid-cell (or region); and FC, TTW, and POI stand for fully connected, time-traffic-weather, and point-of-interest, respectively.}
    \label{fig:dap_model}
\end{figure}

To better utilize heterogeneous sources of data and perform real-time traffic accident prediction, we propose a deep neural network model, named the Deep Accident Prediction (DAP). This model is shown in Figure~\ref{fig:dap_model}, and we describe its components as follows. 

\begin{itemize}[leftmargin=*]
    \item \textbf{Recurrent Component}: Regarding the definition of our prediction framework, we use a set of 8 vectors, each of size 24 (i.e., time-variant attributes), which can be treated as a sequence of such vectors (given their temporal order); therefore, we may benefit from the recurrent neural network models. Specifically, we use a long-short-term-memory (LSTM) model \cite{hochreiter1997long} as represented in Figure~\ref{fig:dap_model}, which includes two recurrent layers, each with 128 LSTM cells. Thus, the output is a vector of size 128. 
    \item \textbf{Embedding Component}: Given the index of a grid-cell, this component provides a distributed representation of that cell which encodes essential information in terms of spatial heterogeneity, traffic characteristics, and impact of other environmental stimuli on accident occurrence. This distributed representation will be derived as we train the entire pipeline. We feed this representation to a feed-forward layer of size 128 that uses the $sigmoid$ activation function. Note that the embedding matrix is of size $|R| \times 128$, where $R$ is the set of all grid-cell regions in input dataset. 
    \item \textbf{Description-to-Vector Component}: This component utilizes the natural language description of historical traffic events in a grid-cell, that is, Desc2Vec data. We feed Desc2Vec of a grid-cell to a feed-forward layer of size 128 using the $sigmoid$ activation function. 
    \item \textbf{Points-of-Interest Component}: This component utilizes points-of-interest data (a vector of size 13), which is a representation of spatial characteristics. We feed a POI vector to a feed-forward layer of size 128 which also uses the $sigmoid$ activation function. 
    \item \textbf{Fully-connected Component}: This component utilizes the output of above components to make the final prediction. Here we have four dense layers of size 512, 256, 64, and 2, respectively. Additionally, to speed-up the training process, we use batch normalization \cite{ioffe2015batch} after the second and the third layer. We use ReLU as the activation function of the first three layers, and apply {\it softmax} on the output of the last layer. 
\end{itemize}

\noindent The DAP model utilizes inputs of various types to better capture temporal and spatial heterogeneity. Using DAP we are able to extract latent spatio-temporal features in terms of embedding representations, whose impact we show through our real-world experiments. We employed grid-search to perform hyper-parameter tuning to find the optimal number of recurrent layers (choices of $\{1,2,3\}$); the best type of recurrent cells (choices of $\{Vanilla\text{-}RNN,\, GRU,\, LSTM\}$); size of the embedding vector for grid-cells (choices of $\{50, 100, 150\}$); sizes of the different fully connected layers (choices of $\{64, 128, 256,$ $512\}$); and activation function for each fully connected layer (choices of $\{sigmoid,\, ReLU,\, tanh\}$). We employed the Adam optimizer \cite{kingma2014adam} with an initial learning rate of $0.01$ to train the model. 
\section{Experiments and Results}
\label{sec:results}
In this section we first describe the data which is used for prediction and analysis. Then, we describe baseline models. Next we compare different models using a variety of metrics, followed by analyses of data attributes. All implementations are in Python using Tensorflow~\cite{tensorflow2015-whitepaper}, Keras~\cite{chollet2015keras}, and scikit-learn~\cite{scikit-learn} libraries; and experiments were run on nodes at the Ohio Supercomputer Center \cite{OhioSupercomputerCenter1987} \footnote{Code and sample data is available at \url{https://github.com/mhsamavatian/DAP}.}. 

\subsection{Data Description}
\label{subsec:data_description}
\begin{figure}[h]
    \vspace{-5pt}
    \centering
    \small 
    \includegraphics[scale=0.6]{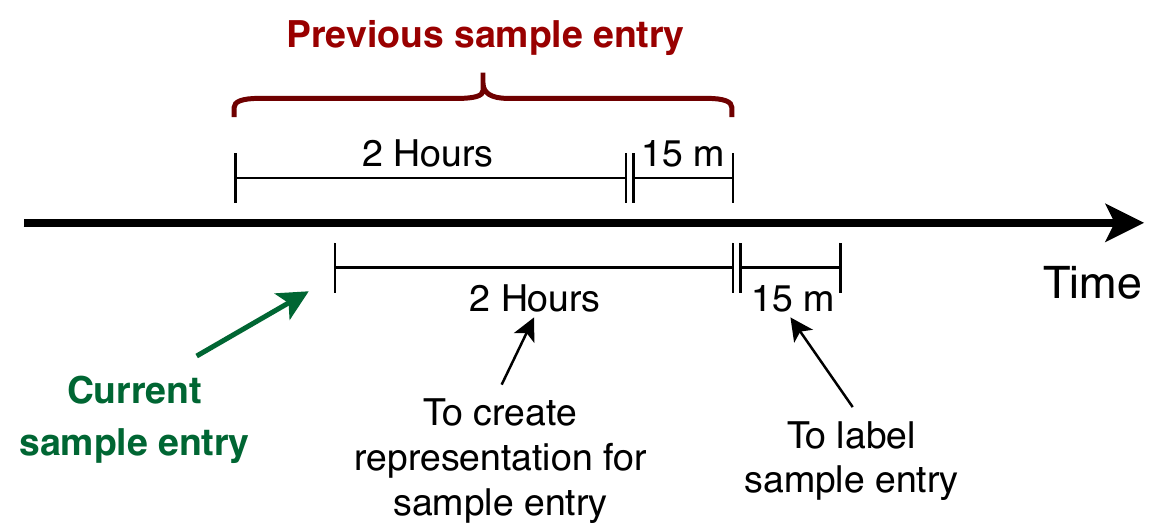}
    \vspace{-5pt}
    \caption{Creating a Sample Entry (see Section~\ref{subsec:data_description}).}
    \label{fig:data_point}
    \vspace{-5pt}
\end{figure}

To evaluate our accident prediction framework, we chose six cities: \textit{Atlanta}, \textit{Austin}, \textit{Charlotte}, \textit{Dallas}, \textit{Houston}, and \textit{Los Angeles}; primarily so as to achieve diversity in traffic and weather conditions, population, population density, and urban characteristics (road-network, prevalence of urban versus highway roads, etc.). We sampled a subset of data (traffic, weather, etc.) collected from June 2018 to August 2018 (i.e., 12 weeks) for each city. We chose this time period to prevent any noises as result of seasonality in weather and traffic patterns. To create Desc2Vec for each grid cell region, we used traffic events which took place within that region from June 2017 to May 2018 (i.e., a one-year time frame), where data obtained from the {\it Large-Scale Traffic and Weather Events dataset} \cite{moosavi2019short}. From the traffic, time, weather, POI, and Desc2Vec data for each grid cell, and by scanning through the data with a window of size 2 hours and 15 minutes and a shift of 15 minutes (see Figure~\ref{fig:data_point}), we built a {\it sample entry} using data of the first two hours (see Section~\ref{subsec:feature_vec}). Each entry is represented by 113 time-invariant and $8 \times 24$ time-variant features. The last 15 minutes is used to label the sample entry as an accident or non-accident case. 

\noindent Since accidents are rare and because our dataset is sparse \footnote{Our data is result of streaming data with possibility of missing records.}, we performed {\em negative sampling} to balance the frequency of samples between accident and non-accident classes. Specifically, we uniformly sampled from the non-accident class with a probability of $2.0\%$. Table~\ref{tab:sample_set} summarizes the number of samples for each class (Acc versus Non-Acc), for each city, after negative sampling. As can be seen, the maximum ratio of accident to non-accident is about $27\%$ for \textit{Los Angeles} (which is still lower than the ratio which is employed by previous studies; e.g., Yuan et al. \cite{yuan2017predicting} employed $33\%$). Table~\ref{tab:sample_set} also shows the number of all other traffic events (except accidents) which took place during the selected 12 weeks time frame. We use data from the first 10 weeks to train and data from the last two weeks as the test set, for each city. 

\begin{table}[ht]
    \vspace{-5pt}
    \small
    \centering
    \setlength\tabcolsep{3pt}
    \caption{\small Distribution of accident (Acc) and non-accident (Non-Acc) classes, and traffic events (except accidents).}
    \vspace{-8pt}
    \label{tab:sample_set}
    \begin{tabular}{lccccc}
        \toprule
        \textbf{City} &&  \#Acc &  \#Non-Acc & Acc/Non-Acc & \#Traffic Events\\
        \midrule
        Atlanta (GA)     &&      2,630 &        11,970     & 22\%    & 24,396\\
        Austin (TX)      &&      4,274 &        23,280     & 18\%    & 16,313\\
        Charlotte (NC)   &&      5,295 &        20,192     & 26\%    & 14,030\\
        Dallas (TX)      &&      3,363 &        28,537     & 12\%    & 28,098\\
        Houston (TX)     &&      5,859 &        43,762     & 13\%    & 40,735\\
        Los Angeles (CA) &&      7,974 &        29,020     & 27\%    & 97,090\\
        \bottomrule
    \end{tabular}
    \vspace{-10pt}
\end{table}
\subsection{Baseline Models}
We chose logistic regression (LR), gradient boosting classifier (GBC), and a Deep Neural Network (DNN) model as baselines.  
\begin{itemize}[leftmargin=*]
    \item [\textbf{-}] \textbf{Logistic Regression (LR)}: A significant number of previous studies leveraged regression-based models to perform accident prediction \cite{chang2005analysis,chen2016learning,chen2018sdcae}. Therefore, we employ logistic regression as a reasonable baseline to perform our binary classification task. 
    \item [\textbf{-}] \textbf{Gradient Boosting Classifier (GBC)}: GBC is a popular general-purpose classification model, with useful boosting characteristics and a suitable learning process. In practice, GBC usually provides superior results for binary or multi-class classification tasks, when compared to the other models such as Random Forest or Support Vector Machine; our preliminary experiments also confirmed this. 
    \item [\textbf{-}] \textbf{Deep Neural Network (DNN)}: This is a four-layer feed-forward neural network, with three hidden layers of size 512, 256, and 64, respectively. $ReLU$ was used as the activation function of the hidden layers, and {\it softmax} was applied on the output of the last layer. To speed-up the training process, we used batch normalization \cite{ioffe2015batch} after the second and third hidden layers. We employed the Adam optimizer \cite{kingma2014adam} with an initial learning rate of $0.01$ to train this model. 
\end{itemize}
As input, the baseline models utilize vectors of size 305, that includes 113 time-invariant and 192 time-variant attributes (see Section~\ref{subsec:feature_vec}). The output is the prediction probability for ``accident'' and ``non-accident'' classes. Using grid-search over heuristic choices of parameters, we found the best parameter setting for each model. For LR, we performed the grid search over choices of regularizations: $\{L1,\, L2\}$, maximum iterations: $\{100,100,10000,100000\}$, and solvers: $\{newton\text{-}cg,\, lbfgs,\, sag,\, liblinear\}$. For GBC, the grid search was performed over choices of learning rates: $\{0.01, 0.05, 0.1,$ $0.15 \}$, number of estimators: $\{100, 200, 300, 400\}$, and maximum depth: $\{3, 4, 5, 6\}$. For DNN, the grid search was performed over choices of initial learning rates: $\{0.001, 0.01, 0.05, 0.1\}$, activation functions: $\{sigmoid, ReLU\}$, number of hidden layers: $\{2, 3, 4\}$, and size of hidden layers: $\{128, 256, 512\}$. 
\begin{table*}[t]
\small
    \centering
    \setlength\tabcolsep{2pt}
    \caption{\small Accident prediction results based on F1-score for class of accidents (Acc), non-accidents (Non-Acc), and weighted average (W-avg).}
    \vspace{-8pt}
    \label{tab:comparing_models}
    \begin{tabular}{l|badc badc badc badc badc}
    \toprule
    \multirow{ 2}{*}{\backslashbox{\textbf{City}}{\textbf{Model}}} & \multicolumn{4}{c}{LR} & \multicolumn{4}{c}{GBC} & \multicolumn{4}{c}{DNN} & \multicolumn{4}{c}{DAP-NoEmbed} & \multicolumn{4}{c}{DAP} \\
     &   Acc & Non-Acc & {W-Avg}& &   Acc & Non-Acc & {W-Avg}& &   Acc & Non-Acc & {W-Avg}& &         Acc & Non-Acc & {W-Avg}& &   Acc & Non-Acc & {W-Avg} \\
    \toprule
    Atlanta    &  0.54 &   0.91 &  0.83 &     &  0.57 &   0.91 &  0.84 &     &  0.62 &   0.89 &  0.83 &     &        0.62 &   0.91 &  0.84 &     &  \textbf{0.65} &   0.89 &  0.84 &     \\
    Austin     &  0.58 &   0.93 &  0.87 &     &  0.61 &   0.93 &  0.87 &     &  0.62 &   0.92 &  0.87 &     &        0.62 &   0.93 &  0.87 &     &  \textbf{0.64} &   0.91 &  0.87 &     \\
    Charlotte  &  0.56 &   0.91 &  0.83 &     &   0.60 &   0.91 &  0.84 &     &  0.61 &   0.87 &  0.82 &     &        0.61 &   0.87 &  0.81 &     &  \textbf{0.63} &   0.87 &  0.82 &     \\
    Dallas     &   0.30 &   0.94 &  0.87 &     &  0.32 &   0.94 &  0.87 &     &  0.36 &   0.94 &  0.87 &     &        0.43 &   0.88 &  0.83 &     &   \textbf{0.50} &   0.93 &  0.88 &     \\
    Houston    &  0.49 &   0.94 &  0.88 &     &  0.51 &   0.94 &  0.88 &     &  \textbf{0.59} &   0.93 &  0.88 &     &        0.58 &   0.92 &  0.88 &     &  0.58 &   0.93 &  0.88 &     \\
    Los Angeles &  0.41 &   0.88 &  0.78 &     &  0.45 &   0.88 &  0.79 &     &  0.53 &   0.81 &  0.75 &     &        0.53 &   0.77 &  0.72 &     &  \textbf{0.56} &   0.84 &  0.78 &     \\
    \bottomrule
\end{tabular}
\end{table*}

\subsection{Exploring Models}
In this section we evaluate different models based on their ability to predict traffic accidents. That is, we compare different models based on $F1\text{-}score$ (defined by Equation~\ref{eq:f1}), reported  for each class separately, as well as the {\it weighted average} $F1\text{-}score$ (the relative frequency of each class is used as its weight). 

\begin{equation}
    \small
    \label{eq:f1}
    \begin{split}
        Precision & = \frac{\text{true positive}}{\text{true positive} + \text{false positive}} \\
        Recall & = \frac{\text{true positive}}{\text{true positive} + \text{false negative}} \\
        F1\text{-}Score & = \frac{2 \times \text{Precision} \times \text{Recall}}{\text{Precision} + \text{Recall}}
    \end{split}
\end{equation}

We used logistic regression (LR), gradient boosting classifier (GBC), and a deep neural network (DNN) model as baselines. We report the result of our DAP model, as well as a variation of DAP without the embedding component (DAP-NoEmbed). We ran each model three times, and reported the average results. As mentioned before, we used grid search to find the optimal parameters. For LR and GBC, we performed this for each city, but for the neural-network-based models we employed grid search for one city and used the best architecture setting for the other cities. DNN, DAP, and DAP-NoEmbed were trained for 60 epochs, and using {\em early stopping} based on the validation set (i.e., $10\%$ of the training set), we used the best model for prediction on the test set. It is worth noting that each model is separately trained and tested for each city and we do not train a single model for all cities. Table~\ref{tab:comparing_models} presents the results of this experiment. In this table we report $F1\text{-}score$ for class of accident (Acc), non-accident (Non-Acc), and the weighted average (W-Avg). We note that the class of accident is usually more important, while we seek to provide reasonable results for the other class (non-accident) as well. LR and GBC usually provide better results for non-accident class, and given the frequency of this class, their weighted average score is also reasonably high. However, when considering the accident class, we note that neural-network-based models provide more satisfactory results, where our proposed DAP model provides superior results for 5 of the 6 cities (DNN provided the best result for Houston). Considering the weighted average on $F1\text{-}score$, we note that DAP provides better results when compared to the other neural-network-based models. 

\begin{figure}[ht]
    \centering
    \includegraphics[scale=0.45]{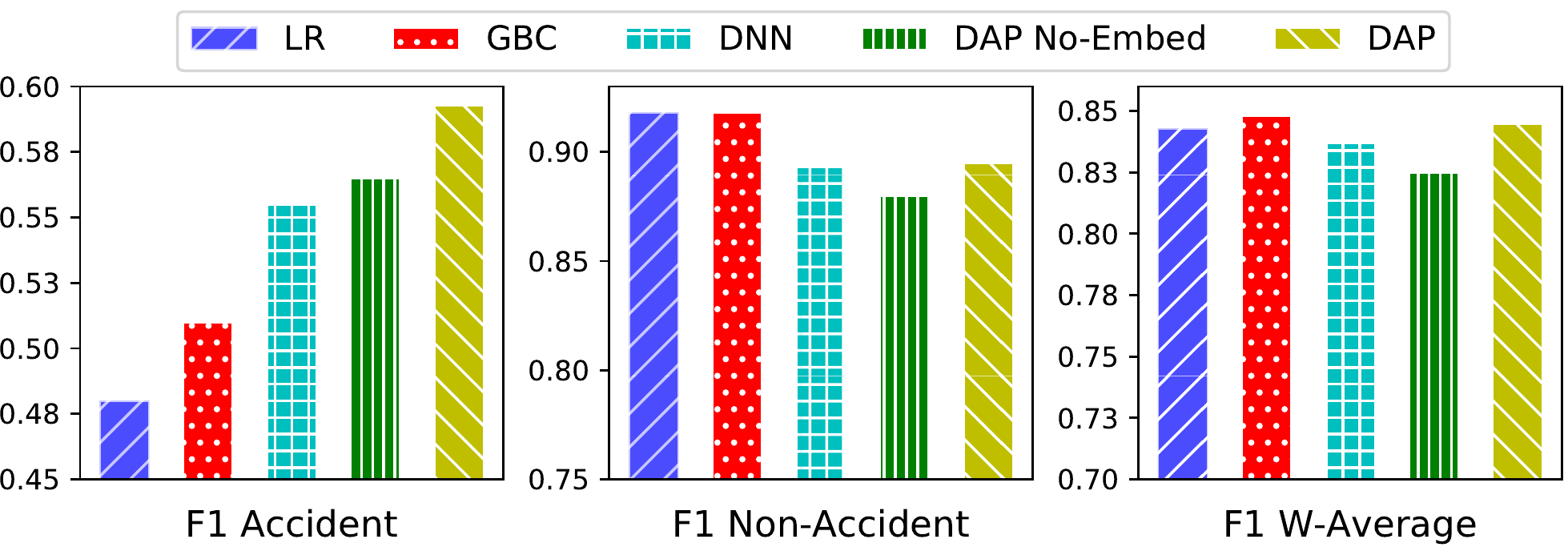}\vspace{-10pt}
    \caption{\small Comparing different models based on average $F1\text{-}score$ (across all six cities) for class of accident, non-accident, and weighted average.}
    \label{fig:dap_vs_all}
\end{figure}

To better compare different models, Figure~\ref{fig:dap_vs_all} shows the average results of different models across all six cities, by separately reporting $F1\text{-}score$ for class of accident and non-accident, and the weighted average $F1\text{-}score$. As one can see, our proposed model provides a significant improvement for class of accidents, while LR and GBC provide slightly better results for the non-accident class. When considering the weighted average, we observe LR, DAP and GBC slightly outperform the other models. Once again note that the ``accident class'' is the one of most importance, given that accidents are rare events. Hence we should pay more attention to false negatives (i.e., predicting an accident as a non-accident) rather than false positives  (i.e., predicting a non-accident as an accident). 

While we cannot directly compare our proposal with the state-of-the-art models such as \cite{yuan2017predicting,chen2016learning,lin2015novel} (due to inconsistency between input types, unavailability of input data used by those models, inconsistency between reported metrics, etc.), we note that their reported results based on $F1\text{-}score$ show similar trend and values (see \cite{yuan2017predicting} for example). Further, we believe that separately reporting prediction results for different classes (i.e., accident versus non-accident) provides a better context to compare different solutions. 
\begin{figure*}[t]
    \small
    \centering
    \hbox{\hspace{-6em}\includegraphics[width=1.2\textwidth]{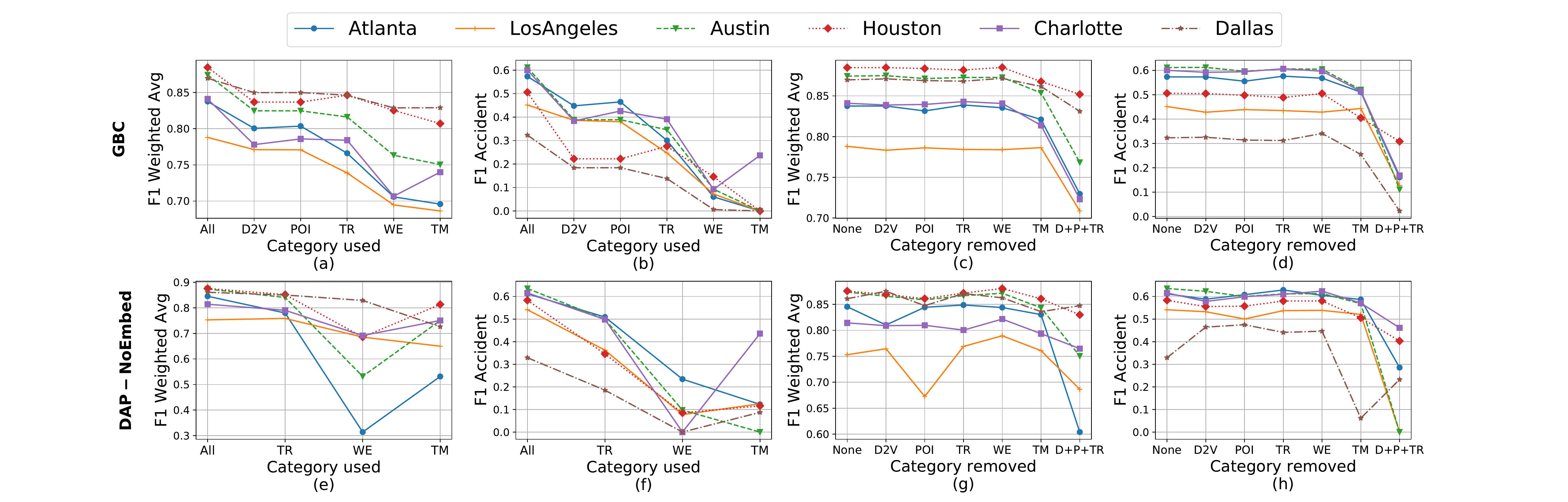}\vspace{-5pt}}
    \caption{\small Prediction results using only one category (a, b, e, and f), and all but one category (c, d, g, and h). Here D2V, TR, WE, and TM stand for Desc2Vec, traffic, weather, and time, respectively. Also, ``D+P+TR'' means removing Desc2Vec, POI, and traffic from input features.}
    \label{fig:feature_analysis}
    \vspace{-5pt}
\end{figure*}

\subsection{Exploring Features}
Our next experiment was to examine the importance of different feature categories for the task of accident prediction. For this exploration, we designed two testing scenarios as follows: 
\begin{itemize}[leftmargin=*]
    \item \textbf{Only One}: This scenario means we only use one category of features (traffic, POI, time, etc.) to perform accident prediction. 
    \item \textbf{All But One}: This scenario means to remove only one category of features and perform the prediction task. 
\end{itemize}
For this experiment, we only report the result of GBC and DAP-NoEmbed, and omit the results of other models for the interest of space. Also, because of having the trainable embedding component, we choose DAP-NoEmbed over DAP to exclude the effect of the embedding component when studying the impact of other features\footnote{Since DAP utilizes an embedding component, we cannot fairly study the impact of several categories of features (in isolation), such as traffic, weather, and points-of-interest; given the correlation between these categories and the latent representation which will be derived for each region.}. 
Figure~\ref{fig:feature_analysis} demonstrates the results, where we report weighted average $F1\text{-}score$, and $F1\text{-}score$ on accident class. Based on parts (a), (b), (e), and (f), we generally observe weather (WE) and time (TM) are the least important categories of attributes to be used alone\footnote{Note that we could not use categories D2V and POI for DAP-NoEmbed, regarding the architecture of this model.}. However, parts (c), (d), (g), and (h) reveal that removing time attributes would significantly hurt the prediction performance. Based on these figures, when we remove Desc2Vec, POI, and Traffic attributes (i.e., D+P+TR), the prediction performance drops significantly, which shows the importance of these categories. We may also note that these categories might have correlation, where removing one of them does not significantly change the prediction results (see (c) and (d)). Therefore, when we remove all three, then we observe a significant drop. 

It is worth noting that among the POI types, we found ``crossing'', ``junction'', ``stop'', and ``traffic signal'' to be more effective than the others for the task of accident prediction. 

\vspace{10pt}
\section{Conclusion and Future Work}
\label{sec:conclusion}
Traffic accidents are a major public safety issue, with much research devoted to analysis and prediction of these rare events. However, most of the studies suffer from using small-scale datasets, relying on extensive data that is not easily accessible to other researchers, and being not applicable for real-time purposes. To address these challenges, we introduced a new framework for real-time traffic accident prediction based on easy-to-obtain, but sparse data. Our prediction model incorporated several neural network based components that used a variety of data attributes such as traffic events, weather data, points-of-interest, and time information. We also created a publicly available countrywide traffic accident dataset, named US-Accidents, through a comprehensive process of data collection, cleansing, and augmentation. Using the data from US-Accidents, we compared our work against several neural-network-based and traditional machine leaning models, and showed its superiority by means of extensive experiments. Further, we studied the impact of different categories of data attributes for traffic accident prediction, and found time, traffic events, and points-of-interest as having significant value. In the future, we plan to incorporate other publicly available sources of data (e.g., demographic information and annual traffic reports) for the task of real-time traffic accident prediction. 

\section*{Acknowledgments}
This work is supported by a grant from the NSF (EAR-1520870), one from the Nationwide Mutual Insurance (GRT00053368), and another from the Ohio Supercomputer Center (PAS0536). Any findings and opinions are those of the authors. 

\vspace{10pt}
\bibliographystyle{ACM-Reference-Format}
\bibliography{main}

\end{document}